\documentclass[pdflatex,sn-mathphys-num]{sn-jnl}% Math and Physical Sciences Numbered Reference Style
\usepackage{graphicx}%
\usepackage{multirow}%
\usepackage{amsmath,amssymb,amsfonts}%
\usepackage{amsthm}%
\usepackage{mathrsfs}%
\usepackage[title]{appendix}%
\usepackage{xcolor}%
\usepackage{textcomp}%
\usepackage{manyfoot}%
\usepackage{booktabs}%
\usepackage{algorithm}%
\usepackage{algorithmicx}%
\usepackage{algpseudocode}%
\usepackage{listings}%
\usepackage{array}
\usepackage{longtable}
\usepackage{caption}
\usepackage{subcaption}

\theoremstyle{thmstyleone}%
\theoremstyle{thmstyletwo}%

\theoremstyle{thmstylethree}%

\begin{document}

\title[Article Title]{MS-SSE-Net: A Multi-Scale Spatial Squeeze-and-Excitation Network for Structural Damage Detection in Civil and Geotechnical Engineering}


\author*[1,4]{\fnm{Saif ur Rehman} \sur{Khan}}\email{saif\_ur\_rehman.khan@dfki.de}
\author[1]{\fnm{Imad Ahmed} \sur{Waqar}}\email{maadiworks@gmail.com}
\author[1]{\fnm{Arooj} \sur{Zaib}}\email{arooj.zaib@dfki.de}
\author[1]{\fnm{Saad} \sur{Ahmed}}\email{saadahmed.waqar@rptu.de}
\author[1,2,3]{\fnm{Sebastian} \sur{Vollmer}}\email{sebastian.vollmer@dfki.de}
\author[1,2,3]{\fnm{Andreas} \sur{Dengel}}\email{andreas.dengel@dfki.de}
\author*[1,2,3,4,5]{\fnm{Muhammad Nabeel} \sur{Asim}}\email{muhammad\_nabeel.asim@dfki.de}

\affil[1]{\orgdiv{Department of Computer Science}, \orgname{Rhineland-Palatinate Technical University of
Kaiserslautern-Landau}, orgaddress{\city{Kaiserslautern}, \postcode{67663}, \country{Germany}}}

\affil[2]{\orgname{German Research Center for Artificial Intelligence}, \orgaddress{\city{Kaiserslautern}, \postcode{67663}, \country{Germany}}}

\affil[3]{\orgname{Intelligentx GmbH (intelligentx.com)}, \orgaddress{\city{Kaiserslautern}, \country{Germany}}}

\affil[4]{\orgname{BiogentX (biogentx.com}, \orgaddress{\city{Renala Khurd, District Okara, Punjab}, \country{Pakistan}}}

\affil[5]{\orgname{Department of Core Informatics, Graduate School of Informatics ,Osaka Metropolitan University}, \orgaddress{\city{Saka, 599-8531}, \country{Japan}}}

\abstract{
Structural damage detection is essential for maintaining the safety and reliability of civil infrastructure. However, accurately identifying different types of structural damage from images remains challenging due to variations in damage patterns and environmental conditions. To address these challenges, this paper proposes \textbf{MS-SSE-Net}, a novel deep learning (DL) framework for structural damage classification. The proposed model is built upon the DenseNet201 backbone and integrates novel multi-scale feature extraction with channel and spatial attention mechanisms (MS-SSE-Net). Specifically, parallel depthwise convolutions capture both local and contextual features, while squeeze-and-excitation style channel attention and spatial attention emphasize informative regions and suppress irrelevant noise. The refined features are then processed through global average pooling and a fully connected classification layer to generate the final predictions. Experiments are conducted on the StructDamage dataset containing multiple structural damage categories. The proposed \textbf{MS-SSE-Net} demonstrates superior performance compared with the baseline DenseNet201 and other comparative approaches. Specifically, the proposed method achieves \textbf{99.31\% precision}, \textbf{99.25\% recall}, \textbf{99.27\% F1-score}, and \textbf{99.26\% accuracy}, outperforming the baseline model which achieved 98.62\% precision, 98.53\% recall, 98.58\% F1-score, and 98.53\% accuracy. These results highlight the effectiveness of integrating multi-scale feature learning with channel spatial attention to enhance discriminative feature representation and improve structural damage classification performance. Beyond structural inspection, the proposed framework also has potential applications in geoscience and engineering fields, such as automated crack detection in geological rock formations and monitoring surface fractures in slopes, tunnels, and underground structures for early hazard assessment and infrastructure maintenance.}
 
\maketitle

\section{Introduction}\label{sec1}
The structural durability of civil infrastructure, such as bridges and dams, is a fundamental baseline for socioeconomic stability, yet aging assets are increasingly compromised by cyclic loading and material degradation. For concrete bridges, specifically, factors such as structural overloading, material deterioration, and long-term creep and fatigue necessitate rigorous monitoring to maintain operational integrity \cite{zhen2025crack}. 
While surface cracking serves as a primary diagnostic indicator for structural health monitoring (SHM), traditional manual inspections are inherently limited by subjectivity, human error, and operational inefficiency \cite{lin2025concrete}. Consequently, the engineering sector is pivoting toward autonomous, high-throughput SHM frameworks leveraged by digital image processing \cite{zhen2025crack} and Artificial Intelligence based techniques \cite{mir2022machine}. This transition is further facilitated by high-resolution data acquisition from Unmanned Aerial Vehicles (UAVs) for bridges \cite{kompanets2025loss} and Closed-Circuit Television (CCTV) systems for dams \cite{ishfaque2026towards}. Albeit used for different infrastructures, these sources provide high-fidelity data essential for training robust DL models.

The integration of DL and computer vision, catalyzed by the adoption of Convolutional Neural Networks (CNNs) and foundation models, has fundamentally reshaped crack detection by enabling automated feature extraction across diverse materials, from modern reinforced concrete \cite{laxman2023automated} to to stone, brick, and masonry structures \cite{zhang2025deep}. Within standard CNN-based frameworks, crack analysis is generally performed across three hierarchical levels: image classification for presence detection, object detection for spatial localization, and semantic segmentation for pixel-level characterization \cite{anusha2025crack}.

The practical efficacy of these approaches has been demonstrated in complex hydraulic structures where manual access is restricted. For instance, Zhang et al. \cite{zhang2023unifying} proposed the UTCD-Net model to leverage transformer architectures, significantly enhancing the detection of multi-scale fractures in rock-fill dams. Similarly, Tang et al. \cite{tang2023novel} introduced a specialized backbone refinement algorithm that demonstrated superior accuracy in quantifying crack geometries in reservoir environments. These advancements validate the potential for high-precision, automated engineering applications in large-scale infrastructure \cite{tang2023novel, zhang2023unifying}. Despite these successes, the implementation of granular defect analysis presents significant challenges. While classification and object detection provide general localization, semantic segmentation is required to delineate precise fracture topologies. However, the efficacy of segmentation is heavily constrained by the requirement for extensive pixel-wise annotated datasets. This labeling process is notoriously labor-intensive and computationally demanding, representing a critical bottleneck that limits the real-world scalability of highly accurate CNN models in SHM \cite{anusha2025crack}.

\subsection{Problem formulation and novelty}
\subsubsection*{Problem formulation}
The structural durability of civil infrastructure, such as bridges and dams, is essential for ensuring public safety and long-term socioeconomic stability. However, aging infrastructure is increasingly affected by cyclic loading, environmental stress, and material degradation, making continuous SHM a critical requirement. Surface cracks and structural defects often serve as primary visual indicators of deterioration, and traditional manual inspection methods are time consuming, subjective, and prone to human error. Recent advances in digital image processing and DL have enabled automated damage inspection using high-resolution imagery captured by UAVs and CCTV systems. Given an input image $x \in \mathbb{R}^{H \times W \times C}$, the goal is to learn a mapping function 
\[
f_\theta : x \rightarrow y,
\]
where $y \in \{1,2,\dots,K\}$ represents the structural damage class among $K$ categories, and $\theta$ denotes the learnable parameters of the model. CNNs extract hierarchical feature representations $F = f_\theta(x)$ through stacked convolution operations. However, many existing CNN-based approaches rely primarily on single scale convolutional kernels, which limits their ability to simultaneously capture fine-grained crack patterns and broader contextual structural information, such as the crack type. Mathematically, the convolution operation for a feature map $F$ can be expressed as
\[
F_{out} = \sigma (W * F_{in} + b),
\]
where $*$ denotes convolution, $W$ is the learnable kernel, $b$ is the bias, and $\sigma$ represents a nonlinear activation function. When using a single kernel scale, the receptive field becomes restricted, reducing the model’s ability to represent structural defects occurring at multiple spatial scales. Furthermore, conventional architectures often treat all feature channels equally, without explicitly highlighting the most informative structural damage patterns. This limitation reduces discriminative capability when dealing with complex textures and environmental variations. Therefore, an effective SHM model should incorporate both multi-scale feature extraction and adaptive feature reweighting mechanisms. To address these challenges, this study proposes an enhanced architecture based on DenseNet201 that integrates multi-scale feature learning with channel and spatial attention mechanisms, enabling the model to selectively emphasize relevant structural damage features and improve classification performance.
\subsubsection*{Contribution and Novelty}
The main contributions of this work are summarized as follows:
\begin{itemize}

\item We propose \textbf{MS-SSE-Net}, a novel DL framework for structural damage classification that enhances feature representation through multi-scale feature extraction and attention-based feature refinement.

\item We introduce a \textbf{MS-SSE block} that employs parallel depthwise convolutions with different kernel sizes to capture both fine-grained local patterns and broader contextual structural damage features.

\item The proposed MS-SSE block integrates \textbf{channel and spatial attention mechanisms} and is incorporated into the \textbf{DenseNet201 backbone} to improve discriminative feature learning and highlight informative structural damage regions.

\item Extensive experiments on the \textbf{StructDamage dataset} demonstrate that the proposed MS-SSE-Net outperforms the baseline DenseNet201 and comparative models, achieving a classification accuracy of \textbf{99.26\%}.

\end{itemize}

\section{Related Work}\label{sec2}
The evolution of automated crack detection has progressed from simple binary classification to high-precision semantic segmentation capable of quantifying geometric properties. However, robust detection still faces challenges due to imperfections in subject image quality, such as variations in texture \cite{chaiyasarn2022integrated}, illumination, exposure, blurriness \cite{guo2024automatic}, and surface conditions \cite{yu2022vision}.

Recent research has pivoted toward architectures that analyze the unique morphology of cracks. Ruan et al. \cite{ruan2025skpnet} developed SKPNet, a semantic segmentation framework which utilizes Dynamic Snake Convolution (DSC) and a Kolmogorov-Arnold Network (KAN) for crack detection in bridges with complex backgrounds. The model is able to capture crack continuities achieving a mean Intersection over Union (mIoU) of 0.877. While effective, the model exhibits sensitivity to domain shifts, particularly when encountering varied background textures. 
For pavement surfaces, Zhang et al. \cite{zhang2025algorithm} introduced U-Net-FML, an enhanced U-Net variant for pixel-level concrete pavement crack segmentation that integrates lightweight convolutions, feature-map partitioning, multipath propagation, and layer-wise multiscale fusion to quickly handle cracks from complex backgrounds. U-Net-FML reaches an mIoU of 76.4\%, F1-score of 74.2\%, 84.2\% precision and 66.4\% recall. A relatively lower Recall value (66.4\%), suggests that while the model is highly precise in identifying detected cracks, it may still overlook extremely fine or hairline fractures that possess low contrast relative to the surrounding concrete aggregate. In the specialized domain of steel infrastructure, Kompanets et al. \cite{kompanets2025loss} introduced a novel loss function inversion technique within a CNN-based framework. It addressed extreme class imbalance in fatigue crack detection, which significantly reduced false positives in crack-free images and achieved an F1-score of 72\% on the Cracks in Steel Bridges (CSB) dataset.

To enhance diagnostic depth, unified pipelines have been developed to execute multiple computer vision tasks concurrently. Ashraf et al. \cite{ashraf2024crack} proposed a unified model integrating RetinaNet, ResNet-34, and U-Net with a shared ResNet-50 backbone, enhanced by transformer-based attention for pavement crack detection, classification, and segmentation. The model achieves high F1-scores ($\approx$94\%) across various pavement crack types, such as alligator, longitudinal, and transverse cracks. However, the concurrent execution of these multiple architectural heads introduces significant computational overhead, which limits its real-time feasibility. Similarly, Sorilla et al. \cite{sorilla2024uav} adopted a two-stage CNN with transfer learning approach for concrete crack detection and segmentation in architectural structures, they paired AlexNet for initial classification with YOLOv4 for subsequent segmentation, and acheived accuracies of 91.5\% for classification and 91.42\% for segmentation. In terms of classification precision, Matarneh et al. \cite{matarneh2024automatic} demonstrated that fine-tuned transfer learning of the ImageNet pre-trained CNN model EfficientNetB3 could classify asphalt cracks (longitudinal, transverse, alligator) with 96\% accuracy, while Yamaguchi et al. \cite{yamaguchi2024road} improved road crack detection by incorporating deep convolutional autoencoders and self-organizing maps (SOM) in CNNs to cluster non-crack background features, reaching a 96\% accuracy rate.

A critical frontier in SHM is maintaining performance under adverse conditions and on edge devices. Shen et al. \cite{shen2025adaptive} addressed low-light environments through the ALF-ViT model, which uses adaptive filtering (ALF) to enhance images before Transformer-based segmentation, achieving an F1-score of 80.3\% on the Crack500 dataset. However, the high parameter count of such Vision Transformers (ViT) often precludes their use on micro-UAVs. For real-time applications, Gao et al. \cite{gao2025using} enhanced the YOLOv11 architecture to improve building crack detection, targeting vertical, horizontal, multi-level, and intricate crack patterns, particularly addressing challenges with small targets and intricate patterns on building surfaces. The enhanced model achieved an 88.6\% mAP@0.5. Benchmarking studies, such as those by Fan \cite{fan2025evaluation}, have evaluated mainstream algorithms by introducing an improved confusion matrix based on linear features and identified that these algorithms, the Single Shot Detector (SSD) offers a robust balance for concrete inspection, particularly when processing VARI-indexed images, yielding an accuracy of 90.6\%. Table~\ref{tab:related_work} presents 
overview of recent methods for crack detection.
{
\renewcommand{\arraystretch}{1.3} % Adjusted for clean vertical spacing without linespace

\begin{longtable}{p{1.8cm} p{3.2cm} p{3.2cm} p{3.5cm}}
\caption{Summary of recent DL-based crack detection methods}
\label{tab:related_work}\\

\toprule
\textbf{References} & \textbf{Method} & \textbf{Objective} & \textbf{Limitation} \\ 
\midrule
\endfirsthead

\toprule
\textbf{References} & \textbf{Method} & \textbf{Objective} & \textbf{Limitation} \\ 
\midrule
\endhead

\bottomrule
\endfoot

Ruan et al. \cite{ruan2025skpnet}, 2025 & SKPNet: Dynamic Snake Conv (DSC) + Kolmogorov-Arnold Network (KAN) & Ensure topological continuity in bridge crack segmentation & Sensitivity to domain differences; performance decreases on varying backgrounds. \\ 

Zhang et al. \cite{zhang2025algorithm}, 2025 & U-Net-FML: Improved U-Net & Pixel-level pavement crack segmentation in complex backgrounds & Low recall (66.4\%) on extremely fine/hairline cracks. \\ 

Kompanets et al. \cite{kompanets2025loss}, 2025 & 2-stage CNN + loss function inversion & Fatigue crack segmentation in steel bridges & High computational overhead from the 2-stage feedback loop. \\ 

Ashraf et al. \cite{ashraf2024crack}, 2024 & RetinaNet + ResNet + U-Net with transformer attention & Multi-task pavement crack detection and segmentation & Transformer compute overhead; sensitive to unseen conditions. \\ 

Sorilla et al. \cite{sorilla2024uav}, 2024 & 2-Stage CNN (AlexNet + YOLO) & UAV-based concrete crack detection in hard-to-reach areas & Performance degrades in complex outdoor environments. \\ 

Matarneh et al. \cite{matarneh2024automatic}, 2024 & Transfer Learning (EfficientNetB3) & Asphalt pavement crack classification & Classification-only; no segmentation; domain shift risk. \\ 

Yamaguchi et al. \cite{yamaguchi2024road}, 2024 & CNN + Self-Organizing Map (SOM) & Road crack detection by reducing false positives & Primarily classification; compute overhead from SOM pipeline. \\ 

Shen et al. \cite{shen2025adaptive}, 2025 & ALF-ViT: Vision Transformer + adaptive learning filters & Pixel-level concrete crack segmentation (low-light) & High compute; low-light data scarcity; domain shift sensitivity. \\ 

Gao et al. \cite{gao2025using}, 2025 & Improved YOLOv11 & Building crack detection for inspection & Performance drops on very low-contrast surface fractures. \\ 

Fan \cite{fan2025evaluation}, 2025 & Linear Confusion Matrix + SSD & Refined evaluation of DL models for crack shape/linearity & Evaluation-only; limited to linear defects. \\ 

\end{longtable}}

\section{Methodology}\label{sec3}

\subsection{Backbone Model Selection}
To determine an effective backbone architecture for the proposed framework, several state-of-the-art CNN models were evaluated on the StructDamage dataset \cite{ijaz2026structdamage}. These models constitute of several imageNet models. As shown in Table~\ref{tab:BS}, among all evaluated models, \textbf{DenseNet201} achieved the highest performance with an accuracy of \textbf{98.62\%}, precision of \textbf{98.53\%}, recall of \textbf{98.58\%}, and F1-score of \textbf{98.53\%}. 

The superior performance of DenseNet201 can be attributed to its dense connectivity mechanism. Unlike traditional CNN architectures where each layer receives input only from the previous layer, DenseNet establishes direct connections between all layers. Formally, the output of the $l$-th layer in DenseNet can be expressed as:
\[
x_l = H_l([x_0, x_1, x_2, \dots, x_{l-1}]),
\]
where $x_l$ represents the output feature map of the $l$-th layer, $H_l(\cdot)$ denotes a composite nonlinear transformation (such as Batch Normalization, ReLU, and convolution), and $[\cdot]$ represents the concatenation operation. This dense connectivity promotes feature reuse and improves gradient flow during training.

Compared with other architectures such as EfficientNetV2, MobileNet, ResNet, and VGG, this mechanism enables DenseNet201 to learn richer and more discriminative feature representations for structural damage patterns. Therefore, based on its superior empirical performance and effective feature propagation capability, DenseNet201 is selected as the backbone network for the proposed MS-SSE-Net framework.
\begin{table}[h!]
\centering
\caption{Performance comparison of different CNN architectures on the StructDamage dataset}
\label{tab:BS}
\small 
\begin{tabular}{lcccc}
\toprule
\textbf{Method} & \textbf{Accuracy (\%)} & \textbf{Precision (\%)} & \textbf{Recall (\%)} & \textbf{F1-Score (\%)} \\
\midrule
DenseNet121 \cite{huang2017}     & 98.43 & 98.33 & 98.38 & 98.34 \\
DenseNet169 \cite{huang2017}     & 98.43 & 98.29 & 98.33 & 98.30 \\
\textbf{DenseNet201}  \cite{huang2017}    & \textbf{98.62} & \textbf{98.53} & \textbf{98.58} & \textbf{98.53} \\
EfficientNetV2B0 \cite{tan2021efficientnetv2} & 96.27 & 96.06 & 96.05 & 96.00 \\
EfficientNetV2B1 \cite{tan2021efficientnetv2} & 93.91 & 93.75 & 93.82 & 93.67 \\
MobileNetV1  \cite{howard2017mobilenets}    & 97.47 & 97.29 & 97.29 & 97.41 \\
MobileNetV2  \cite{sandler2018mobilenetv2}    & 98.33 & 98.22 & 98.24 & 98.21 \\
ResNet50V2  \cite{he2016identity}     & 98.31 & 98.19 & 98.25 & 98.21 \\
ResNet101V2   \cite{he2016identity}   & 98.33 & 98.22 & 98.23 & 98.22 \\
ResNet152V2   \cite{he2016identity}   & 98.24 & 98.15 & 98.14 & 98.13 \\
VGG16      \cite{simonyan2014very}      & 94.89 & 94.70 & 94.65 & 94.57 \\
VGG19      \cite{simonyan2014very}      & 94.84 & 94.62 & 94.62 & 94.51  \\
\bottomrule
\end{tabular}
\end{table}

\subsubsection*{DenseNet Model}
DenseNet201 \cite{huang2017densely} is a densely connected CNN that promotes feature reuse and mitigates the vanishing gradient problem through direct connections between all layers within each dense block. In this architecture, every layer receives feature maps from all preceding layers as input and passes its own feature maps to all subsequent layers, which enables efficient gradient flow and reduces the total number of parameters compared to traditional residual networks.

\begin{figure}[!ht]
    \centering
    \includegraphics[width=1\linewidth]{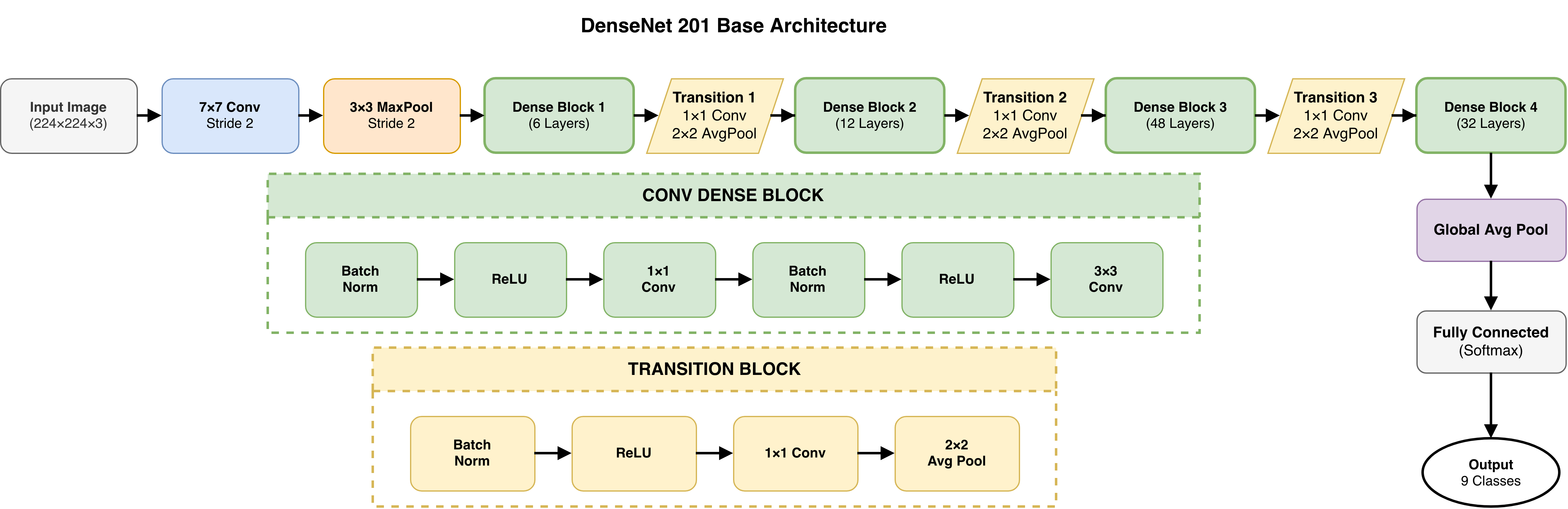}
    \caption{Base architecture of the DenseNet201 backbone used in this study.}
    \label{fig:densenet-architecture}
\end{figure}

The backbone architecture of DenseNet201 is illustrated in Fig.\ref{fig:densenet-architecture}. It consists of an initial $7\times7$ convolution (stride 2) followed by a $3\times3$ max-pooling layer, four dense blocks, and three transition layers. Each dense block comprises multiple $1\times1$ and $3\times3$ convolution units. The transition layers reduce spatial dimensions and feature-map depth via $1\times1$ convolution and $2\times2$ average pooling. The network ends with a global average pooling layer and a softmax classifier. This design allows for high parameter efficiency and the extraction of complex morphological signatures while maintaining a lower memory footprint than comparable deep architectures. Its robust representational power makes it an ideal candidate for image classification and transfer learning applications for crack type detection.

\subsection{Proposed Multi-Scale Spatial-Squeeze Excitation Module}
\begin{figure}[!ht]
    \centering
    \includegraphics[width=0.7\linewidth]{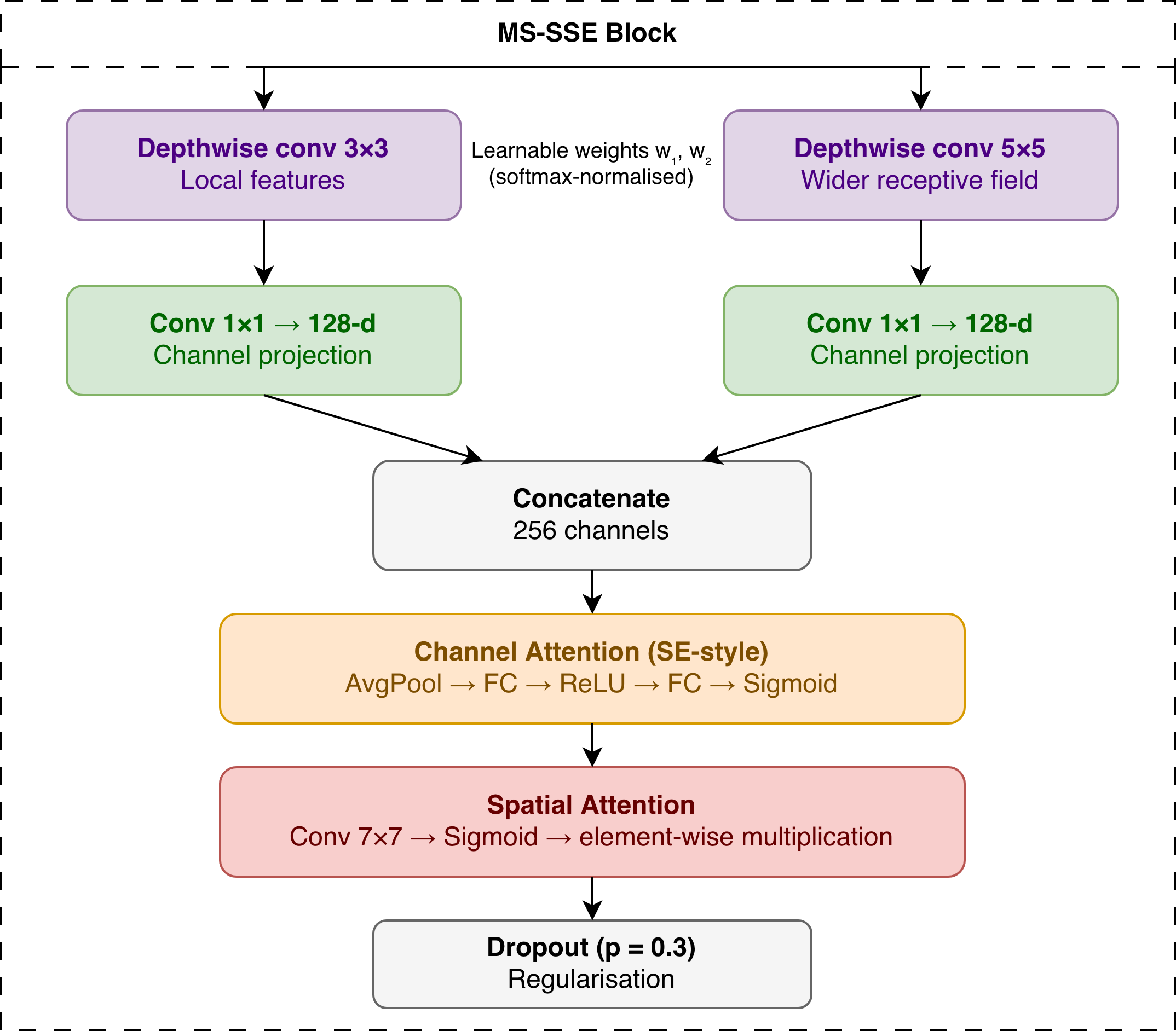}
    \caption{Architecture overview of the proposed MS-SSE block}
    \label{fig:ms-sse-architecture}
\end{figure}

Building upon the DenseNet201 backbone described above, the proposed model introduces a novel MS-SSE module (Fig.\ref{fig:ms-sse-architecture}) that enhances feature representation by jointly capturing multi-scale spatial information and applying hybrid channel-spatial attention. As illustrated in Fig. \ref{fig:ms-sse-architecture}, the module takes the 1920-channel feature maps (7×7 spatial resolution) from the final layer of the frozen DenseNet201 backbone and processes them through two parallel depthwise convolutional branches, a $3\times3$ depthwise convolution for local fine-grained features and a $5\times5$ depthwise convolution for a wider receptive field. Each branch is followed by a $1\times1$ convolution that projects the features to 128 dimensions, after which the outputs are concatenated to form a 256-channel representation.

This enriched feature map then passes through a channel attention block (SE-style squeeze-and-excitation) followed by a spatial attention block implemented as a $7\times7$ convolution with sigmoid activation and element-wise multiplication. A dropout layer (p = 0.3) is applied for regularization, and a residual connection is added to preserve the original information flow. Global average pooling produces a compact 256-dimensional feature vector, which is fed into a fully connected layer and a softmax classifier for the nine class crack type detection task.

The MS-SSE module is lightweight yet highly effective, as it explicitly models both multi-scale spatial context and inter channel relationships while maintaining gradient flow through the residual link. This design enables the model to better discriminate crack patterns and damage types across heterogeneous structural materials compared to standard attention mechanisms.

\subsection{Overview of the Proposed Architecture}
\begin{figure}[!ht]
    \centering
    \includegraphics[width=0.9\linewidth]{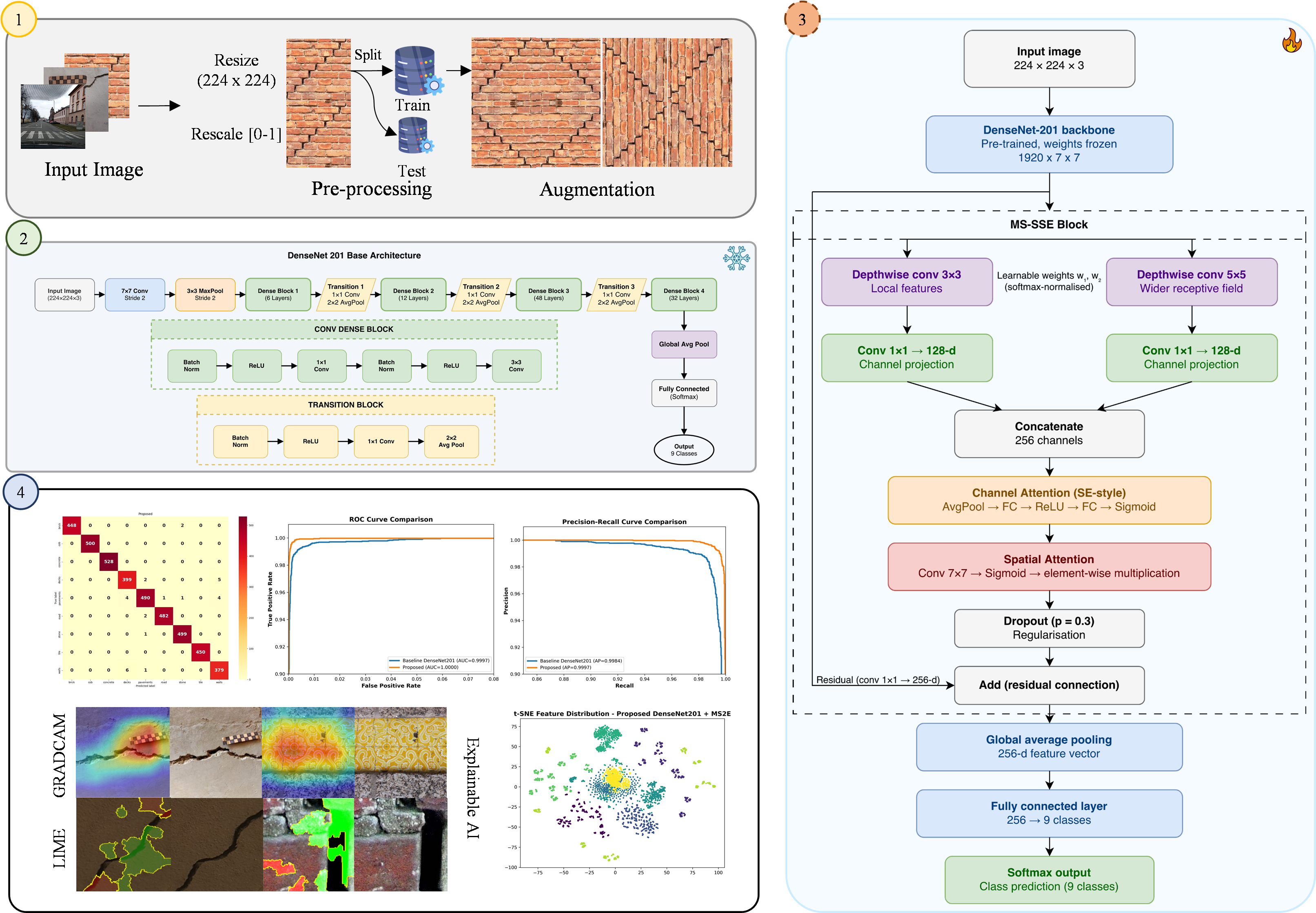}
    \caption{Overview of proposed methodology framework}
    \label{fig:overviewPM}
\end{figure}

The proposed \textbf{MS-SSE-Net} is designed to improve structural damage classification by integrating multi-scale feature extraction with attention-based feature refinement (Fig.\ref{fig:overviewPM}). The framework is built upon the \textbf{DenseNet201} backbone, which is employed for its strong feature propagation capability and dense connectivity mechanism. The DenseNet201 network first extracts hierarchical feature representations from the input structural damage images. The input images are obtained from the StructDamage \cite{ijaz2026structdamage} dataset, which contains diverse crack patterns across multiple structural materials and environmental conditions. To reduce class imbalance and redundancy within the dataset, similarity-based subsampling using perceptual hashing was applied to majority classes, while minority classes were enhanced through data augmentation. These preprocessing steps ensure balanced and diverse training samples, enabling the proposed architecture to learn robust structural damage representations.

To further enhance the discriminative capability of the extracted features, a novel MS-SSE block is introduced. This block incorporates parallel depthwise convolution operations with different kernel sizes to capture both fine-grained local patterns and broader contextual structural features. The multi-scale feature maps are then fused and refined using channel attention and spatial attention mechanisms, which enable the network to emphasize the most informative structural damage regions while suppressing irrelevant background information. After attention-based feature refinement, the enhanced feature representations are passed through a global average pooling layer followed by a fully connected classification layer to generate the final damage class prediction. The architecture enables efficient feature reuse, improved multi-scale representation learning, and adaptive feature weighting, which collectively contribute to improved structural damage classification performance.
\subsection{Hyperparameter configuration and model evaluation metrics}
We trained the architecture using a batch size of 64 for 30 iterations (epochs), with a learning rate of $10^{-3}$. The Adam optimizer was selected for its ability to handle sparse gradients and provide faster convergence via adaptive momentum. 

Beyond simple accuracy \ref{eq:accuracy}, the reliability of the proposed model was assessed via precision \ref{eq:precision}, recall \ref{eq:recall}, and F1-score \ref{eq:f1score} to capture a complete picture of its performance across all target classes. In this context, a true positive (TP) denotes a correctly identified structural damage type, while a true negative (TN) signifies the accurate identification of a non target damage type. Conversely, a false positive (FP) represents a false alarm caused by visually similar damage surfaces, complemented by surface noise, such as shadows or stains, and a false negative (FN) signifies a critical miss-classification
of structural damage type. To consolidate these trade-offs into a single objective measure of classification robustness, the F1-score \eqref{eq:f1score} was utilized, which represents the harmonic mean of the model's precision and recall.
Additionally, Cohen’s Kappa ($\kappa$) \eqref{eq:kappa} was calculated to assess the chance-corrected agreement between the predicted and actual labels. Here $p_o$ represents the  proportionate agreement observed (equivalent to accuracy), while $p_e$ denotes the probability of a random agreement that occurs by chance. This metric ensures model robustness and statistical significance across the multiclass distribution.
\begin{equation}
\text{Accuracy} = \frac{TP + TN}{TP + FP + TN + FN}
\label{eq:accuracy}
\end{equation}

\begin{equation}
\text{Precision} = \frac{TP}{TP + FP}
\label{eq:precision}
\end{equation}

\begin{equation}
\text{Recall} = \frac{TP}{TP + FN}
\label{eq:recall}
\end{equation}

\begin{equation}
\text{F1-Score} = 2 \times \frac{\text{Precision} \times \text{Recall}}{\text{Precision} + \text{Recall}}
\label{eq:f1score}
\end{equation}

\begin{equation}
\kappa = \frac{p_o - p_e}{1 - p_e}
\label{eq:kappa}
\end{equation}

\section{Results and Implementation}\label{sec4}

This section presents a comprehensive evaluation of the proposed MS-SSE-Net model for multi material crack detection. In this section, the experimental setup, dataset characteristics, implementation details, and evaluation metrics are described. Quantitative performance is analyzed through class wise metrics, confusion matrices, and threshold independent curves (ROC and Precision-Recall), followed by benchmarking against 16 representative ImageNet baseline architectures. The performance of the novel MS-SSE module is analyzed relative to seven established attention and structural blocks integrated into the DenseNet201 backbone. Interpretability is further investigated via t-SNE feature embeddings, \textit{Lime} and Grad-CAM heatmaps, providing visual insight into the model's discriminative power. Collectively, these analyses demonstrate the superior accuracy, robustness, and interpretability of the proposed model across diverse materials.

\subsection{Experimental Setup}
The proposed framework was developed using the Keras API within a Python 3.10 environment. To facilitate high-throughput tensor operations and minimize training latency, all computational tasks were executed on an NVIDIA RTX 3050 Tesla GPU workstation with 16 GB of RAM. This configuration provided the necessary memory overhead for large-scale multiclass data handling. The StructDamage \cite{ijaz2026structdamage} dataset was partitioned using a stratified 80:10:10 ratio for training, validation, and testing, respectively.
\subsection{Data Description}

\begin{figure}
     \centering
     \includegraphics[width=1\linewidth]{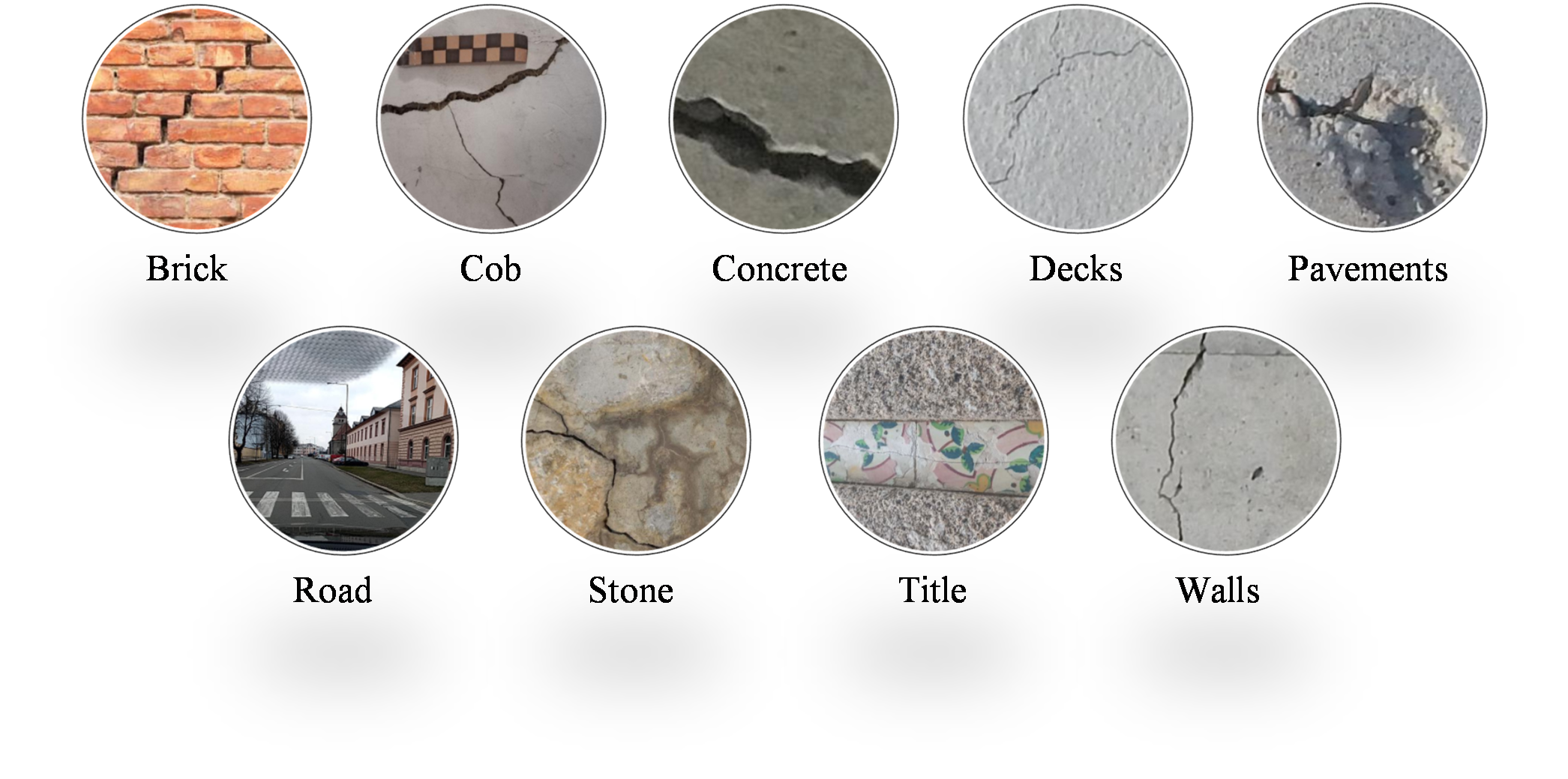}
     \caption{Sample image from each class}
     \label{fig:DS}
 \end{figure}

The StructDamage dataset \cite{ijaz2026structdamage} comprises 78,093 images aggregated from 32 distinct public repositories. These images encompass a wide spectrum of structural materials, environmental conditions, and geographical variations. The dataset is organized into nine categories: brick, cob, concrete, decks, pavements, road, stone, tile, and wall cracks. The class-wise distribution for the full dataset is presented in Table \ref{tab:full-dataset-statistics}. Fig.\ref{fig:DS} illustrates representative sample images from each class. 

\begin{table}[!ht]
\centering
\renewcommand{\arraystretch}{1.3} % More vertical breathing room
\setlength{\tabcolsep}{20pt}      % Adds space between the columns
\caption{Overview of the StructDamage dataset}
\label{tab:full-dataset-statistics}
\begin{tabular}{lr}
\toprule
\textbf{Class} & \textbf{Total Surfaces (images)} \\
\midrule
Brick     & 450    \\
Cob       & 100    \\
Concrete  & 660    \\
Decks     & 2,025  \\
Pavements & 18,404 \\
Road      & 52,518 \\
Stone     & 100    \\
Tile      & 85     \\
Walls     & 3,851  \\
\midrule
\textbf{Total} & \textbf{78,093} \\
\bottomrule
\end{tabular}
\end{table}

\subsubsection{Dataset preprocessing}

A preliminary analysis revealed significant class imbalance within the full StructDamage dataset. This imbalance reflects the varying availability of source imagery rather than a deliberate sampling strategy. Specifically, the Road and Pavements categories dominate the distribution (Table \ref{tab:full-dataset-statistics}), and pose a risk of model bias toward majority classes. To mitigate this, a dual-pronged strategy was employed: similarity-based subsampling for overrepresented classes and data augmentation for minority classes. To eliminate data redundancy, a perceptual similarity filter was applied to majority classes. Each image was assigned a perceptual hash (pHash), and pairwise Hamming distances were calculated. An image was only retained if its similarity index to existing samples was below 30\%. This empirical threshold ensures that the final subset retains visually unique instances and diverse crack morphologies while excluding near-duplicates and repetitive surface textures.

\subsubsection{Augmentation}

\begin{figure}
    \centering
    \includegraphics[width=0.5\linewidth]{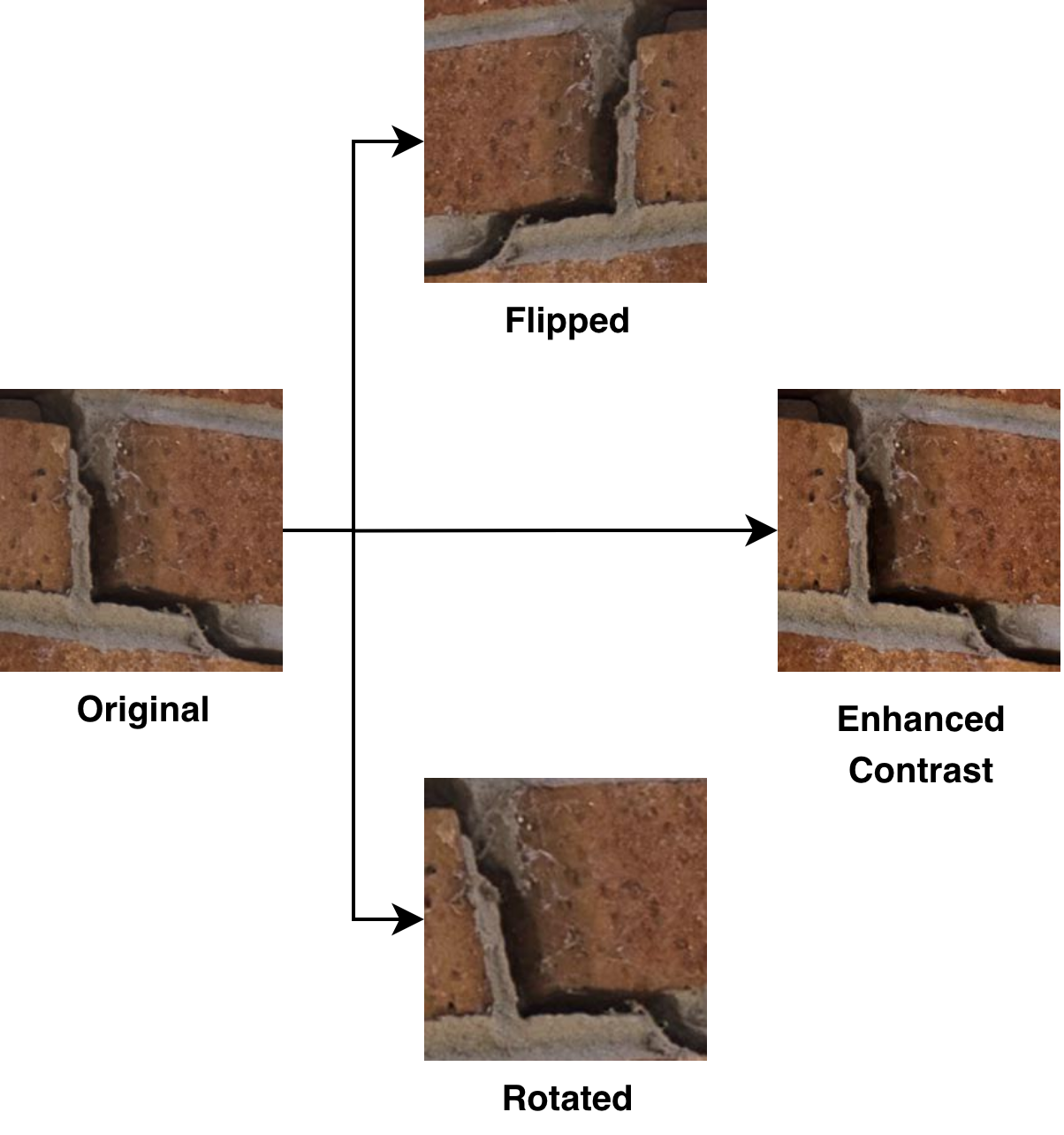}
    \caption{Augmentation sample that shows rotation, flip and contrast enhancement}
    \label{fig:augmentation}
\end{figure}

For minority categories with fewer than 5,000 samples, data diversity was enhanced through geometric and photometric transformations. Geometric adjustments included random horizontal and vertical flips, rotations ($\pm20^\circ$), and minute random cropping. Photometric variations involved brightness and contrast enhancements. To ensure maximum stochastic variation across training epochs without increasing storage requirements, all augmentations were implemented online during training. Samples were generated until each class reached a target threshold of approximately 5,000 images. Fig. \ref{fig:augmentation} illustrates examples of the augmentation steps employed.

\subsubsection{Data Distribution/training procedure for the proposed model}

\begin{figure}
    \centering
    \includegraphics[width=0.8\linewidth]{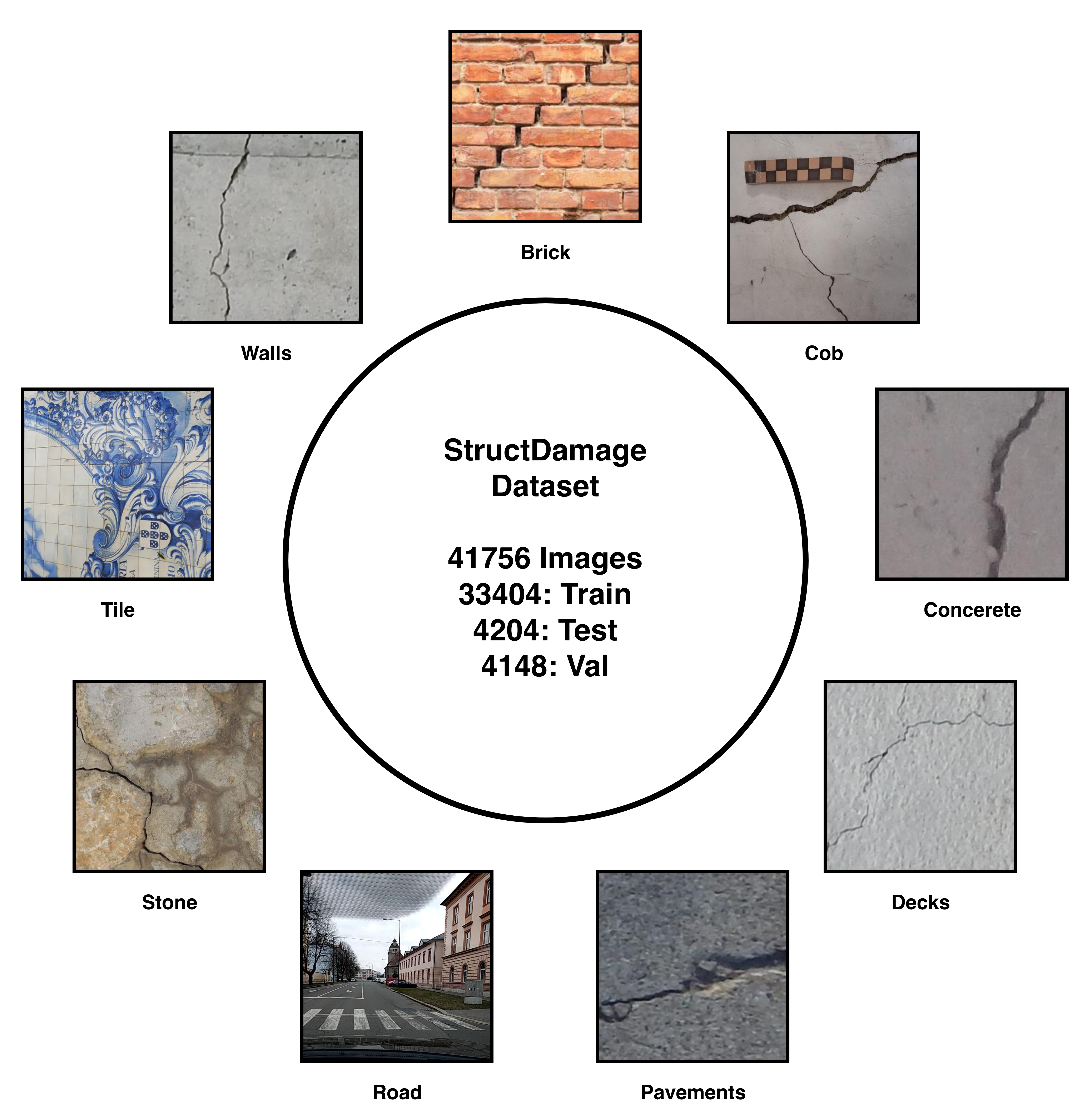}
    \caption{Dataset distribution for training, testing and validation, with sample image from each class}
    \label{fig:Balanced-StructDamage-dataset}
\end{figure}

Following the balancing procedures, the refined corpus consisted of 41,756 images distributed uniformly across the nine categories. This balanced dataset was divided into training (33,404), validation (4,148), and test (4,204) sets using image-level stratification to maintain consistent class proportions (Table \ref{tab:dataset-statistics}). Representative samples from the balanced dataset are shown in Fig. \ref{fig:Balanced-StructDamage-dataset}. While this balanced subset was utilized for the current baseline validation and model training, the full StructDamage dataset (78,093 images) remains the primary release to support varied sampling strategies in future research.

\begin{table}[!ht]
\centering
\small
\caption{Dataset statistics: number of families (distinct surfaces) and sample counts per split}
\label{tab:dataset-statistics}
\begin{tabular}{lrrrr}
\toprule
Class      & Families & Train  & Val   & Test \\
\midrule
Brick      & 450  & 3600 & 450 & 450 \\
Cob        & 100  & 4000 & 500 & 500 \\
Concrete   & 660  & 4224 & 528 & 528 \\
Decks      & 2025 & 3240 & 404 & 406 \\
Pavements  & 4986 & 3997 & 499 & 500 \\
Road       & 4829 & 3863 & 482 & 484 \\
Stone      & 100  & 4000 & 500 & 500 \\
Tile       & 85   & 3400 & 400 & 450 \\
Walls      & 3851 & 3080 & 385 & 386 \\
\midrule
\textbf{Total} & \textbf{13\,086} & \textbf{33\,404} & \textbf{4\,148} & \textbf{4\,204} \\
\bottomrule
\end{tabular}
\end{table}

\subsection{Class-wise performance of evaluation}

To evaluate the results in detail, the selected baseline model, DenseNet201, was first analyzed. As shown in Table \ref{tab:detailed_results}, DenseNet201 achieved a strong overall accuracy of 98.62\% with a Cohen’s Kappa coefficient of 0.9845, demonstrating its effectiveness for structural damage classification. The performance of this backbone can be attributed to its dense connectivity mechanism, which enables efficient feature reuse and improved gradient propagation. However, the class-wise analysis reveals certain limitations. In particular, the precision for the \textit{walls} class dropped to 92.72\%, indicating higher false positive predictions, while recall values for the \textit{pavements} and \textit{decks} classes decreased to 95.40\% and 96.31\%, respectively, suggesting missed detections.

To address these limitations, the proposed MS-SSE-Net extends the DenseNet201 backbone by integrating multi-scale feature extraction with channel and spatial attention mechanisms. As reported in Table \ref{tab:detailed_results}, the proposed model achieves an overall accuracy of 99.31\% and a Cohen’s Kappa coefficient of 0.9922, outperforming the baseline model. At the class level, the model attains perfect precision (100\%) for several categories such as \textit{brick}, \textit{cob}, \textit{concrete}, and \textit{tile}, while maintaining recall values above 98\% across all classes. The macro-average precision, recall, and F1-score reach 99.25\%, 99.27\%, and 99.26\%, respectively, indicates consistent performance across different structural damage categories. These results demonstrate that the integration of multi-scale learning and attention mechanisms significantly enhances the discriminative capability of the DenseNet201 backbone for structural damage classification.

\begin{table}[h!]
\centering
\caption{Class-wise performance metrics for the backbone and proposed model}
\label{tab:detailed_results}
\begin{tabular}{lcccc}
\toprule
\textbf{Model} & \textbf{Class} & \textbf{Precision (\%)} & \textbf{Recall (\%)} & \textbf{F1-Score (\%)} \\
\midrule
DensNet201     & Brick          & 100.00 & 97.78  & 98.88  \\
               &Cob            & 100.00 & 100.00 & 100.00 \\
               &Concrete       & 100.00 & 99.81  & 99.91  \\
               &Decks          & 98.24  & 96.31  & 97.26  \\
               &Pavements      & 98.76  & 95.40  & 97.05  \\
               &Road           & 99.58  & 98.97  & 99.27  \\
               &Stone          & 97.66  & 100.00 & 98.81  \\
               &Tile           & 99.78  & 100.00 & 99.89  \\
               &Walls          & 92.72  & 98.96  & 95.74  \\
\midrule
             &\textbf{Overall Accuracy}    & & &{\textbf{98.62\%}}\\
&Macro Average             & 98.53  & 98.58  & 98.53  \\
&Weighted Average          & 98.66  & 98.62  & 98.63  \\
\bottomrule
\addlinespace[1ex]
\multicolumn{4}{l}{\small \textit{Cohen’s Kappa Coefficient} = 0.9845} \\
\bottomrule

MS-SSE-Net & Brick          & 100.00 & 99.56  & 99.78  \\
           & Cob            & 100.00 & 100.00 & 100.00 \\
           & Concrete       & 100.00 & 100.00 & 100.00 \\
           &Decks          & 97.56  & 98.28  & 97.91  \\
           &Pavements      & 98.79  & 98.00  & 98.39  \\
           &Road           & 99.79  & 99.59  & 99.69  \\
           &Stone          & 99.40  & 99.80  & 99.60  \\
           &Tile           & 100.00 & 100.00 & 100.00 \\
           &Walls          & 97.68  & 98.19  & 97.93  \\
\midrule
& \textbf{Overall Accuracy}           & & & \textbf{99.31\%}   \\
&Macro Average             & 99.25  & 99.27  & 99.26  \\
&Weighted Average          & 99.31  & 99.31  & 99.31  \\
\bottomrule
\addlinespace[1ex]
\multicolumn{4}{l}{\small \textit{Cohen’s Kappa Coefficient }= 0.9922} \\
\bottomrule
\end{tabular}
\end{table}

\begin{figure}[htbp]
\centering
  % Left side: ROC Curve
  \begin{minipage}[b]{0.48\textwidth}
    \centering
    \includegraphics[width=\textwidth]{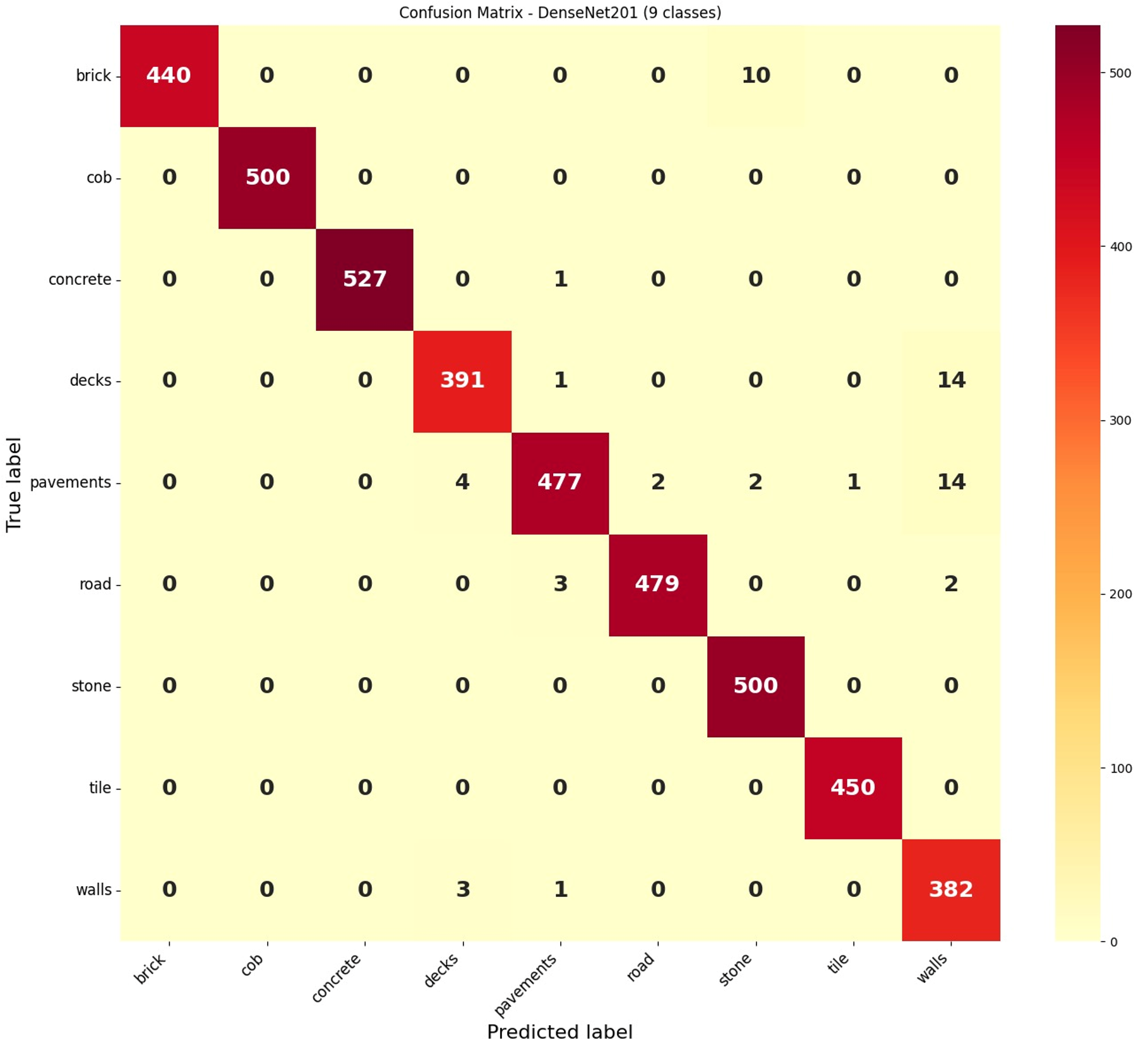}
    \vspace{2pt} % Adjust vertical spacing
    \centerline{\textbf{(a)}}
    \label{fig:confusion-matrix-densenet}
  \end{minipage}
  \hfill
  % Right side: PR Curve
  \begin{minipage}[b]{0.48\textwidth}
    \centering
    \includegraphics[width=\textwidth]{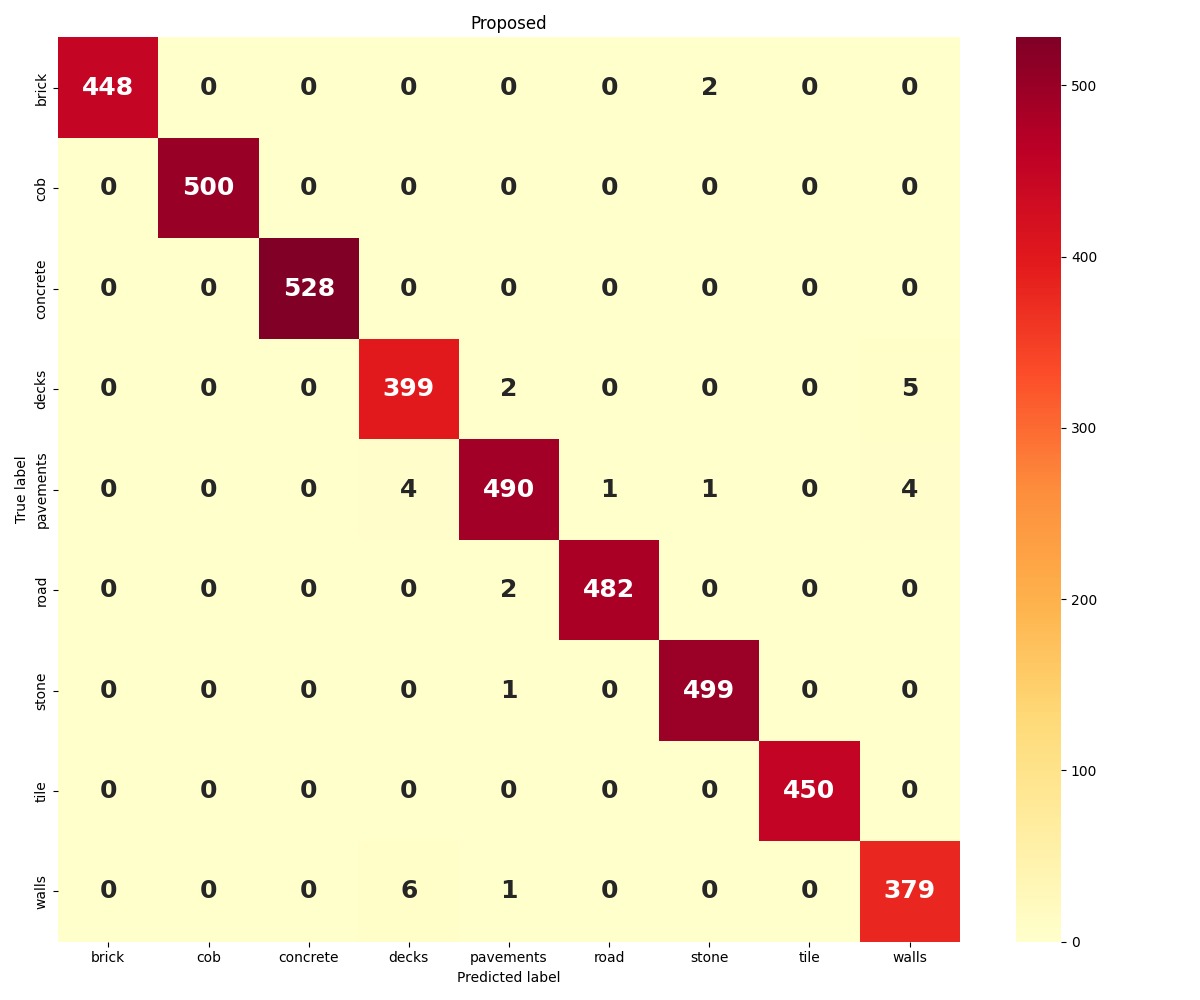}
    \vspace{2pt}
    \centerline{\textbf{(b)}}
    \label{fig:confusion-matrix-proposed}
  \end{minipage}
  \vspace{10pt} % Space between labels and the main caption
  \caption{Performance analysis of baseline and proposed model (a) DenseNet201 (b) proposed (MS-SSE-Net)}
  \label{fig:overall-performance_CM}
\end{figure}

As illustrated by the confusion matrix Fig~\ref{fig:overall-performance_CM} (b), the proposed model demonstrates strong discriminative capability across all nine structural damage categories. Perfect recognition rates were achieved for the cob, concrete, and tile classes, with all corresponding samples correctly classified. Similarly, the brick, stone, and road categories achieved near-perfect accuracies of 99.55\%, 99.80\%, and 99.58\%, respectively, highlighting the model’s ability to learn robust feature representations across diverse structural materials. Minor misclassifications were observed in structurally complex categories such as decks, walls, and pavements, where slight overlaps in crack morphology and surface textures exist. For example, a small number of deck samples were misclassified as walls or pavements, while a few wall instances were confused with decks. These errors are primarily attributed to visual similarities, environmental imaging conditions, and similarities in geometric patterns.

A comparison with the baseline DenseNet201 confusion matrix (Fig~\ref{fig:overall-performance_CM} (a))  further reveals that the backbone model exhibits higher inter-class confusion between visually similar categories. For instance, several decks and pavements samples were incorrectly classified as walls, and some brick samples were misidentified as stone due to textural similarities. Although DenseNet201 achieved strong overall performance, these observations highlight limitations in distinguishing subtle differences in structural patterns. By incorporating multi-scale feature extraction and channel–spatial attention mechanisms on top of the DenseNet201 backbone, the proposed MS-SSE-Net effectively reduces such ambiguities. This improvement is reflected in higher class-wise precision and recall, particularly for challenging categories such as walls and pavements. Consequently, the proposed model achieves a macro-average F1-score of 99.26\%, outperforming DenseNet201 (98.53\%) and demonstrating improved robustness for structural damage classification.

\begin{figure}[htbp]
\centering
  % Left side: ROC Curve
  \begin{minipage}[b]{0.48\textwidth}
    \centering
    \includegraphics[width=\textwidth]{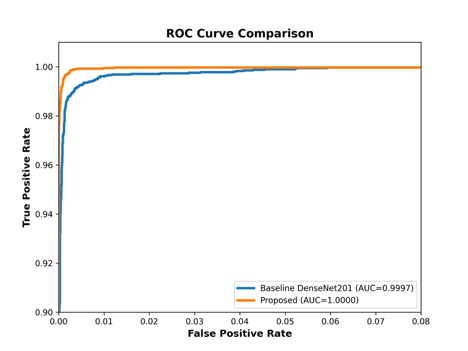}
    \vspace{2pt} % Adjust vertical spacing
    \centerline{\textbf{(a)}}
    \label{fig:AU-ROC-proposed}
  \end{minipage}
  \hfill
  % Right side: PR Curve
  \begin{minipage}[b]{0.48\textwidth}
    \centering
    \includegraphics[width=\textwidth]{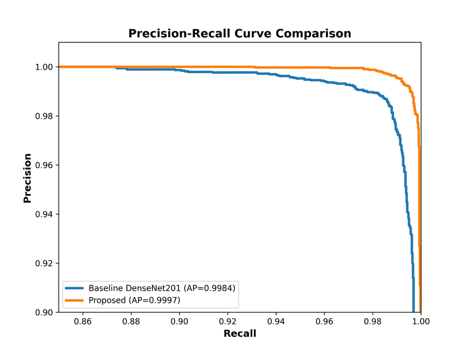}
    \vspace{2pt}
    \centerline{\textbf{(b)}}
    \label{fig:precision-recall-curve}
  \end{minipage}

  \vspace{10pt} % Space between labels and the main caption
  \caption{Performance analysis of proposed model and DenseNet201 (a) AUC-ROC curve (b) Precision-Recall (PR) Curve }
  \label{fig:overall-performance}
\end{figure}

To further validate the discriminatory capability of the proposed MS-SSE-Net architecture, a comparative Receiver Operating Characteristic (ROC) analysis was performed against the strongest baseline DenseNet201. As a threshold-independent performance metric, the ROC curve (Fig. \ref{fig:overall-performance} (a)) illustrates the diagnostic trade-off between the True Positive Rate and the False Positive Rate across the entire range of classification thresholds. As depicted in the comparison, the proposed model exhibits a superior trajectory, approaching the ideal top left coordinate (0,1) with greater acceleration than the DenseNet201 baseline. Specifically, the proposed model achieved a perfect Area Under the Curve (AUC) of 1.0000, marginally outperforming the baseline AUC of 0.9997.

In addition to the ROC analysis, the Precision-Recall (PR) curve was also utilized to evaluate the model performance in the context of positive class identification (Fig. \ref{fig:overall-performance} (b)). While the ROC curve provides a global view of the discriminatory power, the PR curve is a more rigorous indicator of a model's efficacy when dealing with specific class distributions in datasets. The PR curve illustrates the trade-off between Precision and Recall at various classification thresholds, where the goal is to minimize false alarms without avoiding missed detections.

As illustrated in Fig. \ref{fig:overall-performance} (b), the proposed architecture achieved an average precision (AP) of 0.9997, surpassing the already high performance of the DenseNet201 baseline (AP=0.9984). The proposed model’s curve remains exceptionally close to the upper-right corner (1.0,1.0), which indicates that it can maintain a near perfect precision even at maximum recall levels. For practical civil infrastructure inspection, this suggests that the model is highly capable in its capability to identify structural damage with minimal risk of false positives, which can stem from complex and similar surface textures.

\subsection{Performance comparison of MS-SSE-Net with ImageNet models}

To evaluate the comparative superiority of MS-SSE-Net, its performance was benchmarked against sixteen ImageNet trained architectures, which include DenseNet-Base \cite{huang2017densely}, EfficientnetV2 \cite{tan2021efficientnetv2}, ResNetV2 \cite{he2016identity} and ViT \cite{dosovitskiy2020image}. As summarized in Table \ref{tab:baseline_performance_comparison}, the results demonstrate that while several DL baselines achieve high diagnostic efficacy, MS-SSE-Net  consistently outperforms every baseline, attaining 99.31\% accuracy, 99.25\% precision, 99.27\% recall, and 99.26\% F1-score. This represents absolute gains in accuracy of 0.50\% to 5.40\% across the baseline models.

Among the established baselines, DenseNet201 emerged as the most formidable competitor, with an accuracy of 98.62\%. Similarly, the ViT \cite{dosovitskiy2020image} family exhibited strong performance, with the VITB32 and VITL16 variants both achieving accuracies above 98.30\%. Conversely, traditional architectures and lighter weight variants showed slightly diminished reliability. The VGG family (VGG16 and VGG19) and the EfficientNetV2B1 model recorded the lowest performance, with accuracies between 93.91\% and 94.89\%.

Ultimately, MS-SSE-Net outperformed every baseline. It maintains a negligible margin between precision and recall, and shows a highly balanced response that effectively mitigates both false alarms and critical omissions. This consistent $99\%+$ performance across all metrics showcases the model's architectural optimization for structural damage classification, providing a superior balance of robustness and discriminatory power compared to standard ImageNet base models.

\begin{table}[htbp]
\centering
\caption{Comparative performance analysis of MS-SSE-Net and ImageNet models}
\label{tab:baseline_performance_comparison}
\small 
\begin{tabular}{lccccc}
\toprule
\textbf{Category} & \textbf{Method} & \textbf{Accuracy (\%)} & \textbf{Precision (\%)} & \textbf{Recall (\%)} & \textbf{F1-Score (\%)} \\
\midrule
ImageNet & DenseNet121 \cite{huang2017}     & 98.43 & 98.33 & 98.38 & 98.34 \\
& DenseNet169 \cite{huang2017}     & 98.43 & 98.29 & 98.33 & 98.30 \\
& DenseNet201  \cite{huang2017}    & 98.62 & 98.53 & 98.58 & 98.53 \\
& EfficientNetV2B0 \cite{tan2021efficientnetv2} & 96.27 & 96.06 & 96.05 & 96.00 \\
& EfficientNetV2B1 \cite{tan2021efficientnetv2} & 93.91 & 93.75 & 93.82 & 93.67 \\
& MobileNetV2  \cite{sandler2018mobilenetv2}    & 98.33 & 98.22 & 98.24 & 98.21 \\
& ResNet50V2  \cite{he2016identity}     & 98.31 & 98.19 & 98.25 & 98.21 \\
& ResNet101V2   \cite{he2016identity}   & 98.33 & 98.22 & 98.23 & 98.22 \\
& ResNet152V2   \cite{he2016identity}   & 98.24 & 98.15 & 98.14 & 98.13 \\
& VGG16      \cite{simonyan2014very}      & 94.89 & 94.70 & 94.65 & 94.57 \\
& VGG19      \cite{simonyan2014very}      & 94.84 & 94.62 & 94.62 & 94.51\\
Transformer  & VITB16   \cite{dosovitskiy2020image}        & 98.05 & 97.90 & 97.97 & 97.92 \\
& VITB32   \cite{dosovitskiy2020image}        & 98.43 & 98.32 & 98.33 & 98.32 \\
& VITL16   \cite{dosovitskiy2020image}        & 98.33 & 98.22 & 98.24 & 98.22 \\
& VITL32    \cite{dosovitskiy2020image}       & 96.72 & 96.48 & 96.53 & 96.48 \\
\midrule
Our & \textbf{MS-SSE-Net} & \textbf{99.31} & \textbf{99.25} & \textbf{99.27} & \textbf{99.26} \\
\bottomrule
\end{tabular}
\end{table}

\subsection{Performance comparison of novel MS-SSE-Net with existing block}

To validate the architectural benefits of MS-SSE-Net, a comparative analysis was conducted by integrating seven widely used attention mechanisms and structural blocks into the DenseNet201 backbone, which include Channel Attention (CA), Convolutional Block Attention Module (CBAM), Dense Block (DB), Inception Block (NI1), Na\"ive Inception Reduction Block (NI2), Residual Block (RB), and the Squeeze-and-Excitation (SE) Block.

Among the evaluated variants, DenseNet201 + CBAM and DenseNet201 + DB emerged as the most competitive architectures, both achieving a high accuracy of 99.00\% with balanced precision and recall scores exceeding 98.90\%. Similarly, the DenseNet201 + SE variant demonstrated strong diagnostic capability with an accuracy of 98.98\%, while the NI1 and CA modules yielded slightly lower but still respectable accuracies of 98.88\% and 98.57\%, respectively. Conversely, the NI2 block exhibited the lowest performance in this study, with accuracy dropping to 97.65\% and recall falling to 97.51\%. The RB integration also showed a comparative dip in reliability, with an accuracy of 98.22\%.

In contrast, the MS-SSE-Net module achieved the highest scores across all metrics, 99.31\% accuracy, 99.25\% precision, 99.27\% recall, and 99.26\% F1-score. This yields consistent absolute gains of 0.31\% over the best-performing existing block (CBAM/DB) and up to 1.66\% over the weakest variant (NI2). These comparative results provide compelling evidence that the MS-SSE-Net architecture is the key driver behind the overall model superiority reported in prior sections. The enhanced feature recalibration not only boosts aggregate performance but also improves class-wise robustness, as observed by the confusion matrix (Fig~\ref{fig:overall-performance_CM} (b)) and detailed per-class metrics (Table \ref{tab:comparison_study-b}).

\begin{table}[htbp]
\centering
\caption{Comparative analysis of various modules integrated with the DenseNet201 backbone}
\label{tab:comparison_study-b}
\small
\begin{tabular}{lcccc}
\toprule
\textbf{Method} & \textbf{Accuracy (\%)} & \textbf{Precision (\%)} & \textbf{Recall (\%)} & \textbf{F1-Score (\%)} \\
\midrule
CA \cite{woo2018cbam} & 98.57 & 98.48 & 98.45 & 98.46 \\
CBAM \cite{woo2018cbam} & 99.00 & 98.92 & 98.92 & 98.92 \\
DB  \cite{huang2017densenet} & 99.00 & 98.91 & 98.91 & 98.91 \\
NI1 \cite{szegedy2015inception} & 98.88 & 98.87 & 98.75 & 98.80 \\
NI2 \cite{szegedy2016rethinking} & 97.65 & 97.78 & 97.51 & 97.58 \\
RB \cite{he2016resnet} & 98.22 & 98.22 & 98.09 & 98.14 \\
SE \cite{hu2018senet} & 98.98 & 98.91 & 98.86 & 98.89 \\
% \midrule
\textbf{MS-SSE} & \textbf{99.31} & \textbf{99.25} & \textbf{99.27} & \textbf{99.26} \\
\bottomrule
\end{tabular}
\end{table}

\subsection{Performance comparison of MS-SSE-Net with existing studies}
The performance comparison of MS-SSE-Net with existing methods is summarized in Table \ref{tab:performance-comparison}. Our model demonstrates a significant improvement over traditional deep learning architectures. Standard models such as AlexNet (66.98\%) by Yamaguchi et al. \cite{yamaguchi2008image}, VGG-A (70.45\%), and VGG-D (70.61\%) by Simonyan et al. \cite{simonyan2014very} show limited results.
Even complex models like DenseNet-121 (70.77\%) and specialized crack detection frameworks such as YOLO-DEW (83.10\%) and DeeplabV3 (85.00\%) fall short of the precision required for high stakes SHM. Notably, Crack-Net \cite{zhang2017automated}, a domain specific architecture, achieved an accuracy of 87.63\%, yet remains nearly 12\% behind our results. In contrast, MS-SSE-Net achieves a superior accuracy of 99.31\%. Beyond accuracy, while all existing methods were evaluated on smaller datasets, our model is the only one to demonstrate high fidelity performance at a large scale, and proves its robustness for real-world industrial applications.
\begin{table}[ht]
    \centering
    \caption{Performance comparison of MS-SSE-Net with existing methods}
    \label{tab:performance-comparison}
    \begin{tabular}{lllc} 
        \toprule
        \textbf{References} & \textbf{Method} & \textbf{Accuracy} & \textbf{Large-Scale} \\ 
        \midrule
        Yamaguchi et al \cite{yamaguchi2008image}  & AlexNet       & 66.98 & $\times$ \\ % crack
        Simonyan et al. \cite{simonyan2014very}    & VGG-A         & 70.45 & $\times$ \\
        Simonyan et al. \cite{simonyan2014very}    & VGG-D         & 70.61 & $\times$ \\
        Huang et al. \cite{huang2017densely}        & DenseNet-121  & 70.77 & $\times$ \\
        Lu et al. \cite{lu2025deep}      & YOLO-DEW      & 83.10 & $\times$ \\ % crack
        Tang et al. \cite{tang2024cnn}       & DeeplabV3     & 85.00 & $\times$ \\ % crack
        Zhang et al. \cite{zhang2017automated}       & Crack-Net     & 87.63 & $\times$ \\ % crack
        \midrule
        Our                     & MS-SSE-Net      & 99.31 & \checkmark \\
        \bottomrule
    \end{tabular}
\end{table}

\subsection{Industrial application: Interpretable analysis}
Model interpretability helps provide the logical reasoning behind the decision making process. Techniques such as Gradient-weighted Class Activation Mapping (Grad-CAM) and t-Distributed Stochastic Neighbor Embedding (t-SNE) play crucial roles in visualization of the processes which signify the model's prediction abilities. These methodologies allow for a transparent audit of the classification pipeline, showcasing the underlying patterns present in the data. These techniques provide insights into how the model features relevant information that it prioritizes for predictions, help identify potential areas for improvement, aid in transparency and provide trust in its predictions. 

\subsubsection{GRAD-CAM analysis: Visualization of heatmaps for interpretability analysis}
\begin{figure}
    \centering
    \includegraphics[width=0.8\linewidth]{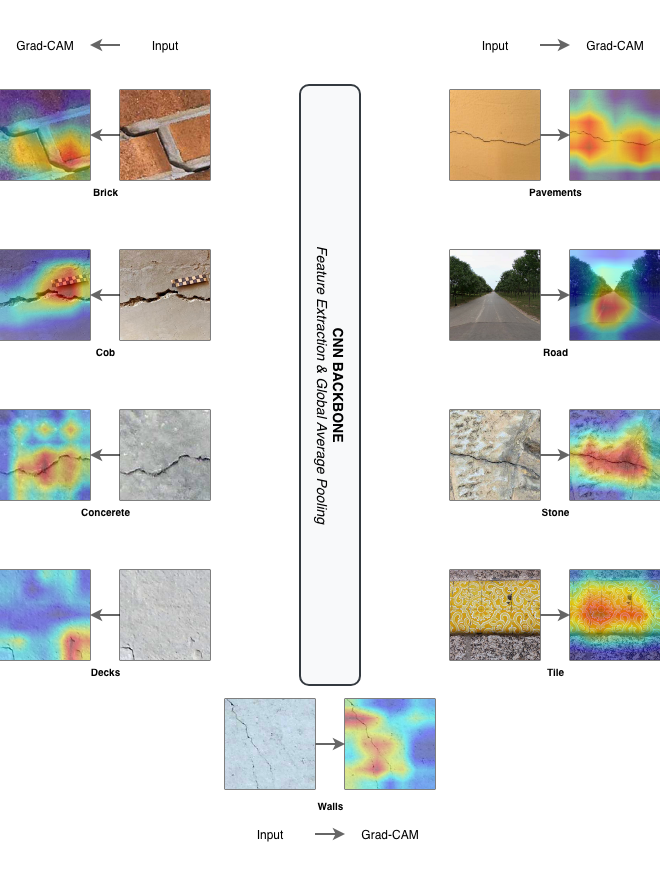}
    \caption{GRADCAM visualization analysis}
    \label{fig:gradcam}
\end{figure}

Fig. \ref{fig:gradcam} displays the Grad-CAM \cite{ijaz2026globalwastedata} visualization for MS-SSE-Net. It was applied to the final convolutional layer of MS-SSE-Net, where it computes the gradients of the predicted class score with respect to the feature maps, and generates intuitive heatmaps that highlight the exact image regions driving each classification decision. The Grad-CAM heatmaps provide a qualitative insight into the model’s internal feature extraction capabilities that accurately identify critical regions of interest.

\subsubsection{LIME interpretability analysis}
\begin{figure}
    \centering
    \includegraphics[width=0.9\linewidth]{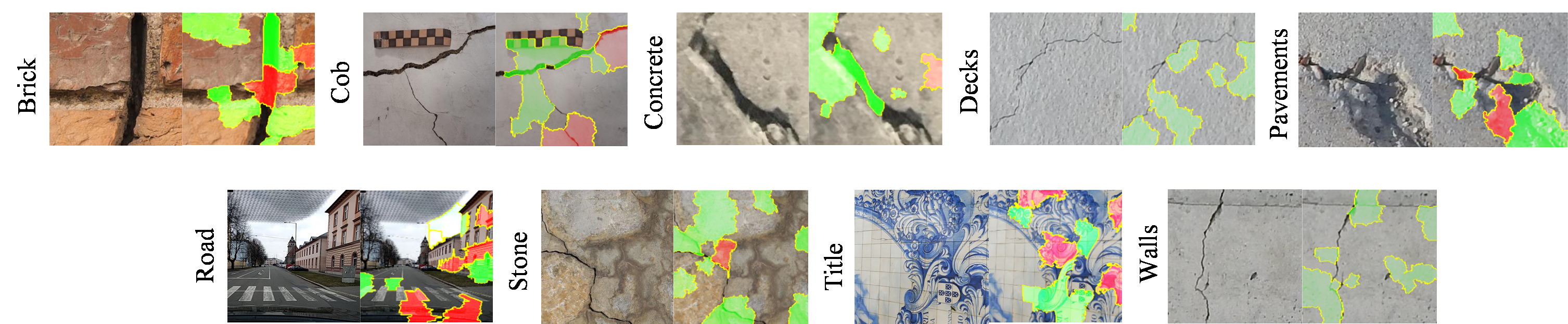}
    \caption{LIME visualization analysis}
    \label{fig:lime}
\end{figure}

To further investigate the decision-making behavior of the proposed MS-SSE-Net, interpretability analysis was conducted using Local Interpretable Model-Agnostic Explanations (LIME) \cite{khan2025dynamic}. LIME provides localized explanations by approximating the model’s complex prediction boundary with a simpler interpretable model around a specific input instance. For each test image, LIME segments the image into superpixels and evaluates the contribution of each region toward the predicted class by perturbing different combinations of these segments. The resulting visualization highlights the most influential regions that contribute positively or negatively to the model classification decision. As illustrated in Fig.~\ref{fig:lime}, the LIME visualizations demonstrate that the proposed MS-SSE-Net consistently focuses on meaningful structural damage regions such as crack edges, fracture intersections, and degraded surface textures across different materials including concrete, masonry, and tiled surfaces. The highlighted superpixels correspond closely with visible crack structures rather than irrelevant background regions such as surrounding textures or lighting variations. This behavior indicates that the multi-scale feature extraction and attention mechanisms within the MS-SSE block effectively guide the network toward structurally relevant patterns. Moreover, the explanations reveal that the model captures both fine crack boundaries and broader contextual damage patterns, validating the effectiveness of the multi-scale design.

\subsection{T-SNE class clustering visualization}

\begin{figure}
    \centering
    \includegraphics[width=1\linewidth]{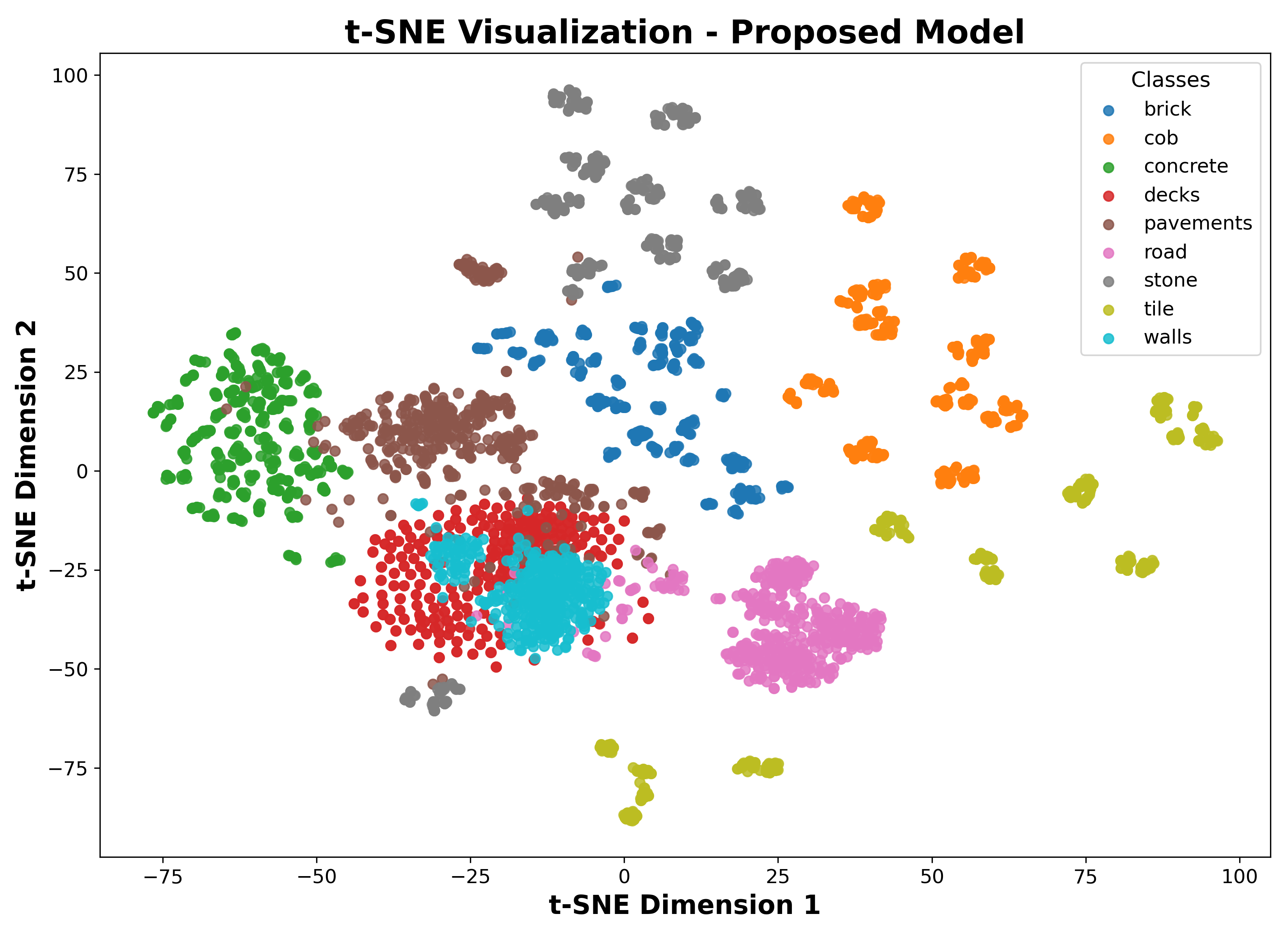}
    \caption{t-SNE visualization of proposed model}
    \label{fig:tsne_proposed}
\end{figure}

To gain qualitative insight into the quality of the learned representations, t-SNE \cite{khan2025dynamic} was performed on the feature vectors extracted from the data by the MS-SSE-Net (Fig.\ref{fig:tsne_proposed}). It is a nonlinear dimensionality reduction technique widely used to visualize high-dimensional data, preserving local structure and pairwise similarities in a low-dimensional space. In the context of classification, t-SNE projections of the feature space provides an intuitive assessment of class separability, where the model's ability to cluster structurally similar instances and separate distinct material categories can be observed.

As illustrated in Fig. \ref{fig:tsne_proposed}, MS-SSE-Net demonstrates adequate feature discriminability, characterized by the formation of dense, well-isolated clusters for the majority of the nine structural classes. This spatial segregation is particularly prominent for categories such as cob, concrete, and tile, which correspond to the perfect 100\% precision and recall scores reported in the classification analysis (Table. \ref{tab:detailed_results}).

An observable inter class overlap occurs between the decks (red) and walls (cyan) clusters in the central region of the embedding. This localized mixing precisely corresponds to the minor confusions identified in the MS-SSE-Net's confusion matrix (Fig. \ref{fig:overall-performance_CM} (b)), which delineated 5 decks misclassified as walls and 6 walls misidentified as decks. Likewise, some overlap can also be seen between pavements (brown), decks (red), and walls (cyan), where 8 pavements have been misidentified as 4 decks and 4 walls respectively. Overall, the pronounced inter-cluster distances and limited cross-talk demonstrate the superior discriminative power of the MS-SSE-Net. 

\begin{figure}
    \centering
    \includegraphics[width=1\linewidth]{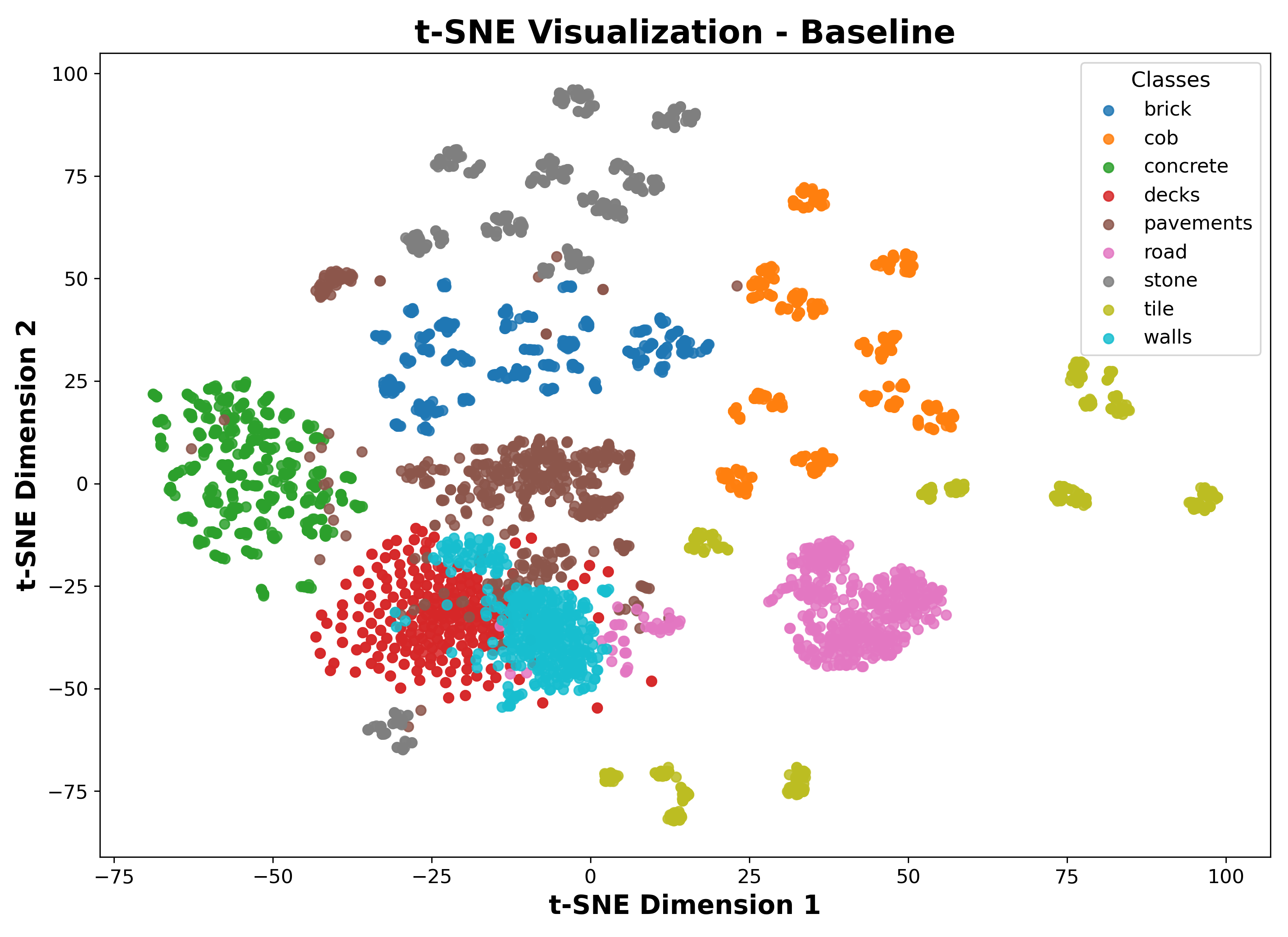}
    \caption{t-SNE visualization of DenseNet201 (backbone) features}
    \label{fig:tsne_baseline}
\end{figure}

To provide a comparative baseline for the feature extraction capabilities of the MS-SSE-Net architecture, a t-SNE visualization was similarly generated for the DenseNet201 model (Fig. \ref{fig:tsne_baseline}). While it demonstrates a clear ability to organize most structural materials into distinct semantic regions, the resulting embedding reveals higher inter-class entanglement compared to MS-SSE-Net. This increased spread in the feature space mirrors the slightly lower overall accuracy (98.62\%) and Cohen’s Kappa (0.9845) recorded for this baseline.

The t-SNE of DenseNet201 is visually similar to that of the MS-SSE-Net, however, with greater amount of misclassifications, due to substantial overlap in the classes present the lower central region of the figure. This overlap exists between the decks (red), walls (cyan), pavements (brown), and to a lesser degree, between brick (blue) and stone (gray). This visual clutter directly corroborates the statistical findings in the DenseNet201 confusion matrix ((Fig~\ref{fig:overall-performance_CM} (a))), where 14 instances of decks and 14 instances of pavements were erroneously mapped to the walls category. Furthermore, the proximity between the brick and stone clusters explains the 10 misclassified samples between these two classes.

\subsection{Ablation study}

To evaluate \cite{maqsood2026automated, hekmat2026fre} the robustness and architectural agnostic nature of the MS-SSE-Net module, a comprehensive series of ablation experiments were conducted. The module was integrated into a wide variety of backbone architectures from
ImageNet pretrained CNNs and Vision Transformer variants, while keeping all other hyperparameters and training conditions identical.

As shown in Tables \ref{tab:cnn_baseline_performance_comparison} and \ref{tab:vit_baseline_performance_comparison}, MS-SSE delivered consistent performance improvements across every tested backbone. Gains ranged from +0.04\% (EfficientNetV2B1) to +0.57\% (ViTB32), with particularly strong enhancements observed on VGG19 (+0.54\%), MobileNetV2 (+0.24\%), ResNet152V2 (+0.24\%), and ViTB16 (+0.31\%), demonstrating near universal effectiveness regardless of whether the underlying architecture relies on spatial locality and translation invariance of CNNs or self attention of transformers.
These results confirm that MS-SSE-Net is not merely complementary to DenseNet201 but provides orthogonal benefits to a broad spectrum of modern CNN and Transformer families. By explicitly modeling multi-scale spatial context and applying hybrid channel-spatial attention, MS-SSE significantly enhances feature discriminability for subtle crack patterns across diverse structural materials. This broad applicability strongly supports that the superior performance of the MS-SSE-Net (99.31\% accuracy) is primarily attributable to the novel MS-SSE design rather than the choice of backbone alone.

\begin{table}[htbp]
\centering
\caption{Comparative performance analysis of finetuned ImageNet models with proposed novel MS-SSE block}
\label{tab:cnn_baseline_performance_comparison}
\small 
\begin{tabular}{lcccc}
\toprule
\textbf{Method} & \textbf{Accuracy (\%)} & \textbf{Precision (\%)} & \textbf{Recall (\%)} & \textbf{F1-Score (\%)} \\
\midrule
DenseNet121 + MS-SSE       & 98.48 & 98.39 & 98.43 & 98.40 \\
DenseNet169 + MS-SSE      & 98.60 & 98.40 & 98.44 & 98.39 \\
EfficientNetV2B0 + MS-SSE & 97.05 & 97.00 & 96.45 & 96.50 \\
EfficientNetV2B1 + MS-SSE & 93.95 & 93.77 & 93.84 & 93.70 \\
MobileNetV2  + MS-SSE     & 98.57 & 98.48 & 98.45 & 98.46 \\
ResNet50V2  + MS-SSE     & 98.36 & 98.29 & 98.32 & 98.31 \\
ResNet101V2 + MS-SSE   & 98.45 & 98.35 & 98.43 & 98.32 \\
ResNet152V2 + MS-SSE  & 98.48 & 98.37 & 98.40 & 98.43 \\
VGG16    + MS-SSE      & 95.15 & 95.10 & 94.77 & 94.87 \\
VGG19    + MS-SSE      & 95.38 & 95.12 & 95.22 & 95.10\\
\bottomrule
\end{tabular}
\end{table}

\begin{table}[htbp]
\centering
\caption{Comparative performance analysis of finetuned transformer models with proposed novel MS-SSE block}
\label{tab:vit_baseline_performance_comparison}
\small 
\begin{tabular}{lcccc}
\toprule
 \textbf{Method} & \textbf{Accuracy (\%)} & \textbf{Precision (\%)} & \textbf{Recall (\%)} & \textbf{F1-Score (\%)} \\
\midrule
VITB16 + MS-SSE          & 98.36 & 98.49 & 98.40 & 98.35 \\
VITB32 + MS-SSE          &99.00 & 98.90 & 98.91 & 98.93 \\
VITL16 + MS-SSE          & 98.48 & 98.33 & 98.42 & 98.46 \\
VITL32  + MS-SSE         & 97.05 & 97.32 & 96.55 & 96.52 \\
\bottomrule
\end{tabular}
\end{table}

% structural damage classification
% MS-SSE-Net

\subsection{Industrial limitations and future work}

The SHM and civil infrastructure inspection industries have long faced two persistent challenges: the scarcity of large, comprehensive, and multi-material crack datasets, and the predominance of single-modality (static image) data. Most publicly available datasets have historically been limited to a single damage type or a narrow set of infrastructure assets, restricting the development of truly generalizable automated systems.

The present work directly addresses these longstanding industrial limitations by introducing the StructDamage \cite{ijaz2026structdamage} dataset, a broad, nine-class collection covering diverse structural materials under varied real world conditions, and by developing a high accuracy classification model on top of it. This combination enables reliable multi-material structural detection and provides a solid foundation for practical deployment.

Nevertheless, certain data driven constraints remain. The current dataset is still single modality (RGB images only) and, while significantly more diverse than prior collections, does not yet encompass video sequences, multi-sensor inputs, or certain rare crack morphologies frequently encountered in operational environments. These data limitations currently restrict the model’s ability to perform temporal tracking, multi modal fusion, or fine grained damage type diagnosis using advanced techniques such as large language models.

To further bridge the gap between laboratory performance and full industrial deployment, future work will focus on expanding the dataset and framework in two directions. The corpus will be extended into a true multi modal collection by incorporating video sequences from UAVs and robotic platforms alongside static images. Additional classes and underrepresented damage morphologies will be included. These enhancements will enable the development of a comprehensive multi-modal system capable of precise, temporally aware damage identification and will support the integration of more advanced diagnostic tools, ultimately delivering a robust end-to-end decision-support solution for real-world infrastructure management.

\section{Conclusion}
This paper presented \textbf{MS-SSE-Net}, a novel architecture designed to enhance structural damage classification. Built upon the DenseNet201 backbone, the proposed model introduces a multi-scale feature extraction strategy combined with channel and spatial attention mechanisms to improve the representation of structural crack patterns across diverse materials and environmental conditions. By integrating depthwise convolutions with multiple kernel sizes and attention-based feature refinement, the model effectively captures both local crack textures and broader contextual structural features. Experiments conducted on the large-scale StructDamage dataset demonstrate that MS-SSE-Net significantly improves classification performance compared to the baseline DenseNet201 and several widely used architectural blocks, including CA, CBAM, RB, DB, and inception-based modules. The proposed MS-SSE-Net achieved an overall accuracy of \textbf{99.31\%} and a macro-average F1-score of \textbf{99.26\%}, outperforming the baseline DenseNet201 across nearly all structural categories. Beyond structural inspection, the proposed framework has broader implications for geoscience and civil engineering applications, where reliable detection of surface fractures, material degradation, and geological cracks is critical for monitoring infrastructure and natural formations. Therefore, MS-SSE-Net can serve as a robust and scalable tool for supporting automated structural health monitoring, infrastructure assessment, and related engineering analysis tasks.

Future work will focus on extending the proposed architecture to real-time deployment in field inspection systems and integrating ensemble data sources. Additionally, exploring lightweight variants of MS-SSE-Net for edge devices and evaluating the model on additional structural and geological datasets will further enhance its applicability in practical engineering environments.
\backmatter

\section*{Declarations}
\begin{itemize}
\item \textbf{Ethics approval and consent to participate.}
\item \textbf{Funding.} No funding
\item \textbf{Declaration of competing interest.} The authors declare that they have no known competing financial interests or personal relationships that could have appeared to influence the work reported in this paper.
\item \textbf{Consent for publication.} Not applicable
\item \textbf{Data availability.} Data available after corresponding request. 
\item \textbf{CRediT authorship contribution statement.} Saif Ur Rehman Khan \& Muhammed Nabeel Asim: Conceptualization, Data curation, Methodology, Software, Validation, Writing original draft \& Formal analysis.Sebastian Vollmer \& Andreas Dengel: Conceptualization, Funding acquisition, Review. Imad Ahmed Waqar, Saad Ahmed \&Arooj Zaib: Software, review \& editing, Data curation.
\end{itemize}

\section*{Supplementary material}
\begin{table}[ht]
\centering
\caption{Acronym and its full form}
\begin{tabular}{ll}
\hline
\textbf{Acronym}  & \textbf{full form}                  \\ \hline
ML & Machine Learning \\
SHM &  Structural Health Monitoring \\
MS-SSE & Multi-Scale Spatial-Squeeze Excitation \\
DL  & Deep Learning   \\
ROC & Receiver Operating Characteristic  \\
AUC     & Area Under the Curve \\
ViT & Vision Transformers  \\
CA & Channel Attention \\
CBAM & Convolutional Block Attention Module \\
DB & Dense Block \\
NI1 &  Inception Block \\
NI2 & Naive Inception Reduction Block \\
RB & Residual Block \\
SE & Squeeze-and-Excitation \\
GRADCAM &  Gradient-weighted Class Activation Mapping \\
t-SNE & t-Distributed Stochastic Neighbor Embedding 
    \\ \hline
\end{tabular}
\end{table}

\noindent

\bibliography{sn-bibliography}% common bib file

\end{document}